\documentclass[10pt,twocolumn,letterpaper]{article}

\usepackage{wacv}              

\usepackage{graphicx}
\usepackage{amsmath}
\usepackage{amssymb}
\usepackage{booktabs}
\usepackage{arydshln}
\usepackage{xcolor}
\usepackage{caption}
\usepackage{multirow}
\usepackage[utf8]{inputenc}
\usepackage[pagebackref,breaklinks,colorlinks]{hyperref}

\usepackage[capitalize]{cleveref}
\crefname{section}{Sec.}{Secs.}
\Crefname{section}{Section}{Sections}
\Crefname{table}{Table}{Tables}
\crefname{table}{Tab.}{Tabs.}

\def\wacvPaperID{838} 
\def\confName{WACV}
\def\confYear{2025}

\begin{document}

\title{Data Augmentation for Surgical Scene Segmentation with Anatomy-Aware Diffusion Models}
\author{
Danush Kumar Venkatesh\textsuperscript{1,2}, Dominik Rivoir\textsuperscript{1,3}, Micha Pfeiffer\textsuperscript{1}, Fiona Kolbinger\textsuperscript{4,5,6},\\
Stefanie Speidel\textsuperscript{1,2,3}
\and
\small{\textsuperscript{1}NCT/UCC Dresden, DKFZ Heidelberg, Faculty of Medicine \& University Hospital Carl Gustav Carus TU Dresden,HZDR Dresden, Germany}\\
\small{\textsuperscript{2}Department of Translational Surgical Oncology, NCT/UCC Dresden, Faculty of Medicine \& University Hospital Carl Gustav Carus Germany}\\
\small{\textsuperscript{3}The Centre for Tactile Internet with Human-in-the-Loop (CeTI), TUD Dresden, Germany} \\
\small{\textsuperscript{4}Weldon School of Biomedical Engineering, Regenstrief Center for Healthcare Engineering (RCHE), Purdue University, USA}\\
\small{\textsuperscript{5}Department of Biostatistics and Health Data Science, Richard M. Fairbanks School of Public Health, Indiana University, USA}\\
\small{\textsuperscript{6}Department of Visceral, Thoracic and Vascular Surgery, University Hospital \& Faculty of Medicine Carl Gustav Carus, TUD Germany}\\
{\tt\small \{firstname.lastname\}@nct-dresden.de, fkolbing@purdue.edu}\\
}
\maketitle

\begin{abstract}

In computer-assisted surgery, automatically recognizing anatomical organs is crucial for understanding the surgical scene and providing intraoperative assistance. While machine learning models can identify such structures, their deployment is hindered by the need for labeled, diverse surgical datasets with anatomical annotations. Labeling multiple classes (i.e., organs) in a surgical scene is time-intensive, requiring medical experts. Although synthetically generated images can enhance segmentation performance, maintaining both organ structure and texture during generation is challenging. We introduce a multi-stage approach using diffusion models to generate multi-class surgical datasets with annotations. Our framework improves anatomy awareness by training organ specific models with an inpainting objective guided by binary segmentation masks. The organs are generated with an inference pipeline using pre-trained ControlNet to maintain the organ structure. The synthetic multi-class datasets are constructed through an image composition step, ensuring structural and textural consistency. This versatile approach allows the generation of multi-class datasets from real binary datasets and simulated surgical masks. We thoroughly evaluate the generated datasets on image quality and downstream segmentation, achieving a $15\%$ improvement in segmentation scores when combined with real images.
\end{abstract}

\section{Introduction}
\label{sec:intro}
\begin{figure*}[ht]
\begin{center}
\includegraphics[width=\textwidth]{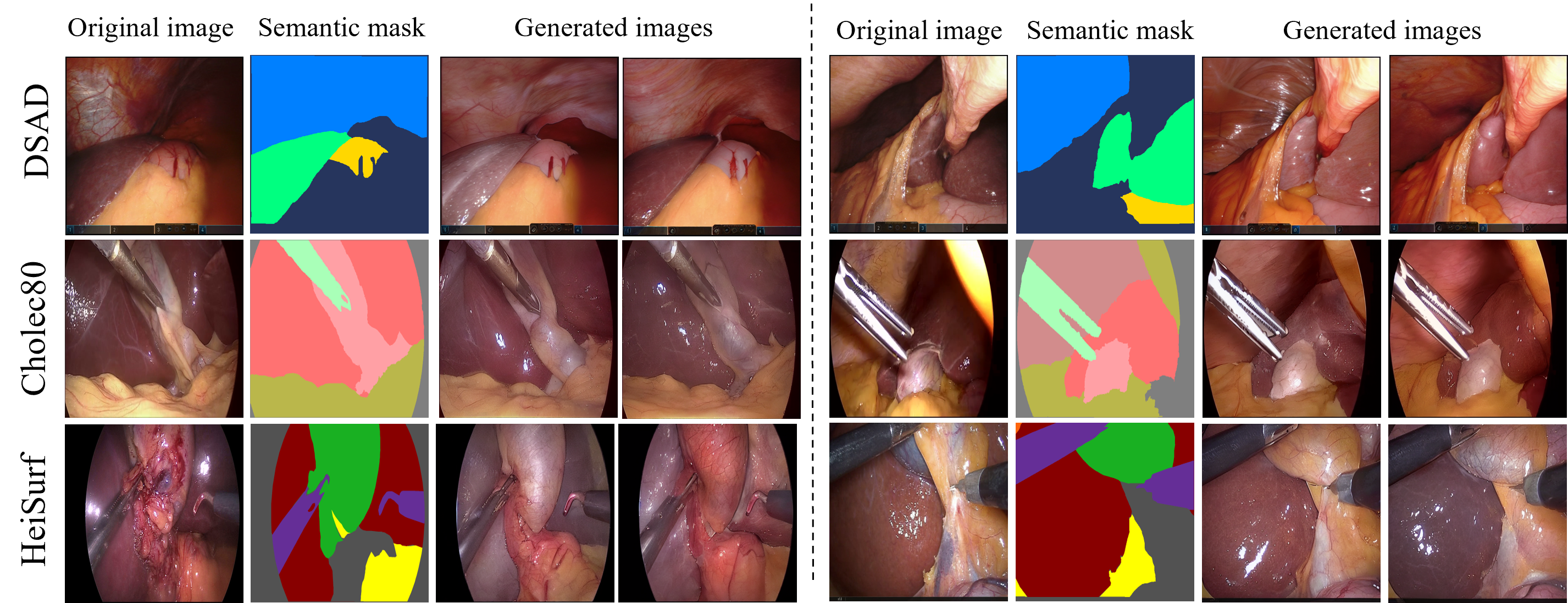}
\end{center}
   \caption{The generated multi-class surgical images (Generated images column) for three different surgical datasets (denoted by name on the left side) with their corresponding semantic masks using our diffusion approach. Our approach can generate realistic and diverse organ textures using the segmentation masks as masking and conditioning signals.}
\label{fig:intro}
\end{figure*}
Computer-assisted surgery (CAS) aims to improve assistance for surgical teams and minimize complications during surgical procedures~\cite{maier2022surgical}. 
A comprehensive understanding of surgical anatomy is essential for implementing such context-aware guidance in laparoscopic surgeries. Moreover, anatomy segmentation is a challenging computer vision task that serves useful in other downstream surgical tasks such as action recognition~\cite{seenivasan2022global}, surgical skill assessment~\cite{vedula2016task,papp2022surgical} and for navigation with segmentation~\cite{docea2024seesaw}. 
Hence, training machine learning models on surgical images to semantically segment each anatomical structure serves as a promising solution to improve interventional healthcare. 

Despite the remarkable progress in deep learning for semantic segmentation, their application to surgical data science faces a hindrance due to the necessity for large-scale diverse and annotated data~\cite{maier2022surgical}. Generating a multi-class dataset requires annotating every pixel representing each anatomical structure within the surgical scene. In contrast, binary annotations pertain to only one subject in an image. The surgical field faces a challenge in that the expertise of medical professionals, i.e., doctors, are required to annotate the datasets, who have limited time resources~\cite{scheikl2020deep}. Consequently, this has led to very limited open-sourced multi-class anatomical datasets such as HeiSurf~\cite{bodenstedt2021heichole} or the Dresden Surgical Anatomy dataset (DSAD)~\cite{carstens2023dresden} with only a few thousand images.  

Simulation environments offer the capability to generate synthetic surgical images. This process is advantageous because different labels, such as semantic masks, depth, and normal maps, can be rendered automatically. Prior works~\cite{chen2019learning,sankaranarayanan2018learning,pfeiffer2019generating,Rivoir_2021_ICCV,venkatesh2024exploring,yoon2022surgical} have demonstrated the usage of synthetic images along with generative models such as generative adversarial networks (GANs)~\cite{goodfellow2014generative} to improve segmentation. 
However, these methods often struggle to generate high-quality images with sufficient diversity. 

Diffusion models (DMs) have emerged as a promising approach to generative modeling~\cite{sohl2015deep,ho2020denoising} and have surpassed the state-of-the-art GAN-based methods in image synthesis~\cite{dhariwal2021diffusion}. For surgical applications, precise control over the shape and structure of organs is essential. However, achieving spatial alignment using only text prompts in DMs is notably challenging~\cite{tumanyan2023plug,parmar2023zero}. Recent research efforts, such as ControlNet~\cite{zhang2023adding} and T2i-Adapters~\cite{mou2023t2iadapter}, have introduced methods to control the output of DMs using additional signals like segmentation masks and edge maps.

In this work, we create synthetic multi-class surgical datasets with full-scene annotations. Our pipeline is based on latent diffusion models conditioned by text prompts. We utilize segmentation masks and formulate a diffusion inpainting objective to make these diffusion models understand the texture properties of different anatomies (anatomy-aware). To precisely generate different anatomical structures, we introduce an inference pipeline consisting of pre-trained ControlNet~\cite{zhang2023adding} with extracted edge images as the controlling signal. We address the challenges of multi-object compositionality by implementing an image-fusing method to produce multi-class surgical images containing diverse anatomical structures. We thoroughly evaluate the generated images and demonstrate their usefulness in the downstream segmentation of different anatomical structures and surgical tools.

A natural question arises: \textbf{What is the available segmentation mask?} We identify two possibilities. Firstly, we have minimal segmentation masks from the open-sourced (OS) surgical datasets (See~\cref{fig:intro}). These masks accurately represent the shapes of various organs in the dataset. We use the OS masks to train the diffusion models and generate synthetic multi-class datasets through inpainting models, resulting in the first synthetic dataset (\emph{Syn}). Secondly, surgical simulations (SS) can provide full-scene segmentation masks, although replicating the exact shapes of organs in laparoscopic setting is time-consuming (See~\cref{fig:syn_comp}). Consequently, the SS masks contain organs which approximate the true shapes similar to those in real datasets. We utilize the SS masks to generate images of different organs, leading to the synthetic dataset (\emph{SS-Syn}) that incorporates the anatomical texture properties from real surgical datasets.

\textbf{Contributions}. We summarize our contributions as follows: 
%
    
\begin{enumerate}
    \item We introduce a multi-stage approach using diffusion models to generate high-quality, realistic surgical images with full annotations in an anatomy-controlled manner. 
    \item By modeling and integrating each semantic class through separate organ-specific diffusion inpainting models, our approach ensures organs' shape and texture consistency in the generated images, surpassing methods that rely on full-scene segmentation masks. An additional benefit of our method is its capacity to generate multi-class datasets using only real binary datasets and multi-class simulation masks. This is especially advantageous when multi-class segmentation labels are limited, but low-cost binary labels are available (see~\cref{res:2}).
    \item We thoroughly evaluate the generated images on image quality and downstream segmentation of organs. Our results show a $15\%$ improvement in segmentation scores when models are trained on combined datasets of our generated and real images. To support research in surgical computer vision, we are releasing the generated data along with their labels at  \url{https://gitlab.com/nct_tso_public/muli-class-image-synthesis}.
\end{enumerate}

\section{Related Work}
\textbf{Diffusion Models} DMs were initially introduced by Sohl-Dickstein et al.~\cite{sohl2015deep} and have recently gained significant attention for their superior image synthesis performance compared to GANs~\cite{dhariwal2021diffusion}. In Latent Diffusion Models (LDMs)~\cite{rombach2022high}, the diffusion process occurs in the latent image space~\cite{esser2021taming}, reducing computational costs. Stable Diffusion (SD)~\cite{rombach2022high} is a large-scale implementation of LDM trained on natural images. The image generation was conditioned using text prompts by encoding the text inputs into latent vectors using pre-trained language models like CLIP~\cite{radford2021learning}. SD remains a competitive open-source LDM model, and we use them to train on our surgical datasets.

\textbf{Controllable Image generation} To enable personalization or customization, controlling the DMs is necessary. Text-based editing has been achieved by adjusting text prompts or manipulating the CLIP features~\cite{avrahami2022blended,brooks2023instructpix2pix,gafni2022make,hertz2022prompt,kawar2023imagic}. 
ControlNet~\cite{zhang2023adding} proposes to attain spatial conditioning via learning adapter networks similar to the Unet in LDMs. Conditioning signals were learned with additional smaller adapters that plug into DMs in T2i-Adapter~\cite{mou2023t2iadapter}. Despite their promise, these methods demand extensive computational resources
and prolonged training times for adaption to the surgical datasets. Additionally, generating multi-subjects (classes) together is still a challenge that has yet to be solved. In our approach, we use the advantages of ControlNet for spatial conditioning on each organ and generate the multi-class(subject) images via a simple image composition step.

\textbf{Surgical image synthesis} Studies on laparoscopic image synthesis have primarily focused on image-to-image (I2I) translation. Computer-simulated surgical images~\cite{pfeiffer2019generating,Rivoir_2021_ICCV,venkatesh2024exploring,yoon2022surgical}, phantom data~\cite{sharan2021mutually} and segmentation maps~\cite{marzullo2021towards} were used with GANs to synthesize realistic surgical images or video data. 
Frisch et al.~\cite{frisch2023synthesising} generated rare cataract surgical images using DDIMs~\cite{song2020denoising} and SD based I2I methods were explored in~\cite{kaleta2024minimal,venkatesh2024surgical}. Our method requires only segmentation masks as input, and hence, this further reduces the simulation modeling efforts. Allmendinger et al.~\cite{allmendinger2023navigating} analyzed diffusion models like Dall-e2~\cite{ramesh2022hierarchical}, Imagen~\cite{saharia2022photorealistic} for laparoscopic image synthesis. Diffusion models have become popular for generating medical data~\cite{kazerouni2023diffusion}, especially MRI~\cite{dorjsembe2022three,khader2022medical} or CT images~\cite{lyu2022conversion}. However, it is essential to note that they differ in imaging modality from surgical data. To our knowledge, this is the first study to focus on generating multi-class surgical image datasets.

\textbf{Segmentation strategy} 
Data augmentation is widely used to amplify existing datasets~\cite{shorten2019survey} and can be easily implemented during training. Both color and geometric augmentations have demonstrated improvements in segmentation performance for medical images~\cite{goceri2023medical}. 
In surgical context, Jenke et al.~\cite{jenke2024model} suggested an implicit labeling method based on mutual information between classes for surgical images. We train segmentation models using various augmentations as baselines and evaluate their performance against our synthetically generated datasets.

\section{Preliminaries}
\subsection{Stable Diffusion}
Diffusion models~\cite{sohl2015deep,ho2020denoising} (DMs) are probabilistic generative models that iteratively generate images by removing noise from an initial Gaussian noise image, $x_{T} \sim \mathcal{N}(0, I)$. This includes a forward process defined as 
\begin{equation}\label{eq:1}
    x_t=\sqrt{\bar{\alpha}_t} x_0+\sqrt{1-\bar{\alpha}_t} z,
\end{equation}
 where $z \sim \mathcal{N}(0, I)$, $\bar{\alpha_{t}}$ is the noise schedule and $x_{0}$ is the clean image. In the backward process
a neural network $\epsilon_{\theta}(x_t, t, P)$ learns to predict the added noise $z_{t}$ at each step $t$ by optimizing the loss, $\mathcal{L}_{DM}$,
\begin{equation}\label{eq:2}
\left.\mathcal{L}_{DM}=\mathbb{E}_{z \sim \mathcal{N}(0,I), t}\left[\| z-\epsilon_\theta\left(x_t, t, P\right)\right) \|_2^2\right],
\end{equation} 
where P is the guiding text prompt. In this work, we leverage the pre-trained text-conditioned Stable Diffusion (SD)~\cite{rombach2022high} model, in which the diffusion process occurs in the latent space using an image autoencoder.

\subsection{ControlNet}
ControlNet (CN) is a framework designed for controlling pre-trained DMs' image generation process by integrating additional conditioning signals such as sketches, key points, edges, and segmentation maps~\cite{zhang2023adding}. The model consists of two sets of U-Net weights derived from the pre-trained DM: with $\theta_{c}$, that undergoes training using task-specific datasets to accommodate the additional condition, $\mathrm{c}_{f}$ and the frozen copy, $\theta_{L}$. 
Let $S_{f}$ be the input feature map from SD, then the feature map $y_{c}$ from the ControlNet is defined as,
\begin{equation}
\boldsymbol{y}_{\mathbb{C}}=\mathcal{B}(S_{f} ; \theta_{L})+\mathcal{C}\left(\mathcal{B}\left(S_{f}+\mathcal{C}\left(\boldsymbol{c_{f}} ; \Theta_{\mathrm{s} 1}\right) ; \theta_{c}\right) ; \Theta_{\mathrm{s} 2}\right),
\end{equation}
where $\mathcal{C}(\cdot;\cdot)$ denotes $1$x$1$ zero-convolution layers with $\Theta_{(\cdot)}$ parameters that links pre-trained SD with ControlNet blocks and $\mathcal{B}(\cdot;\cdot)$ is a neural block with a set of parameters. We use pre-trained CN for spatial conditioning.

\subsection{SDEdit}
SDEdit is an image editing method that uses stochastic differential equations (SDE) to solve the reverse diffusion process~\cite{meng2021sdedit}. A user-given image is firstly noised up to a specific limit depending on the specified noising strength, and denoising starts from this noisy image, which serves as a controlling signal, back to a realistic image. Text prompts can be added as additional guiding signals during the denoising process. This method is used in the final stage for image refinement in our pipeline.

\section{Methodology}
\begin{figure*}
\begin{center}
\includegraphics[width=\textwidth]{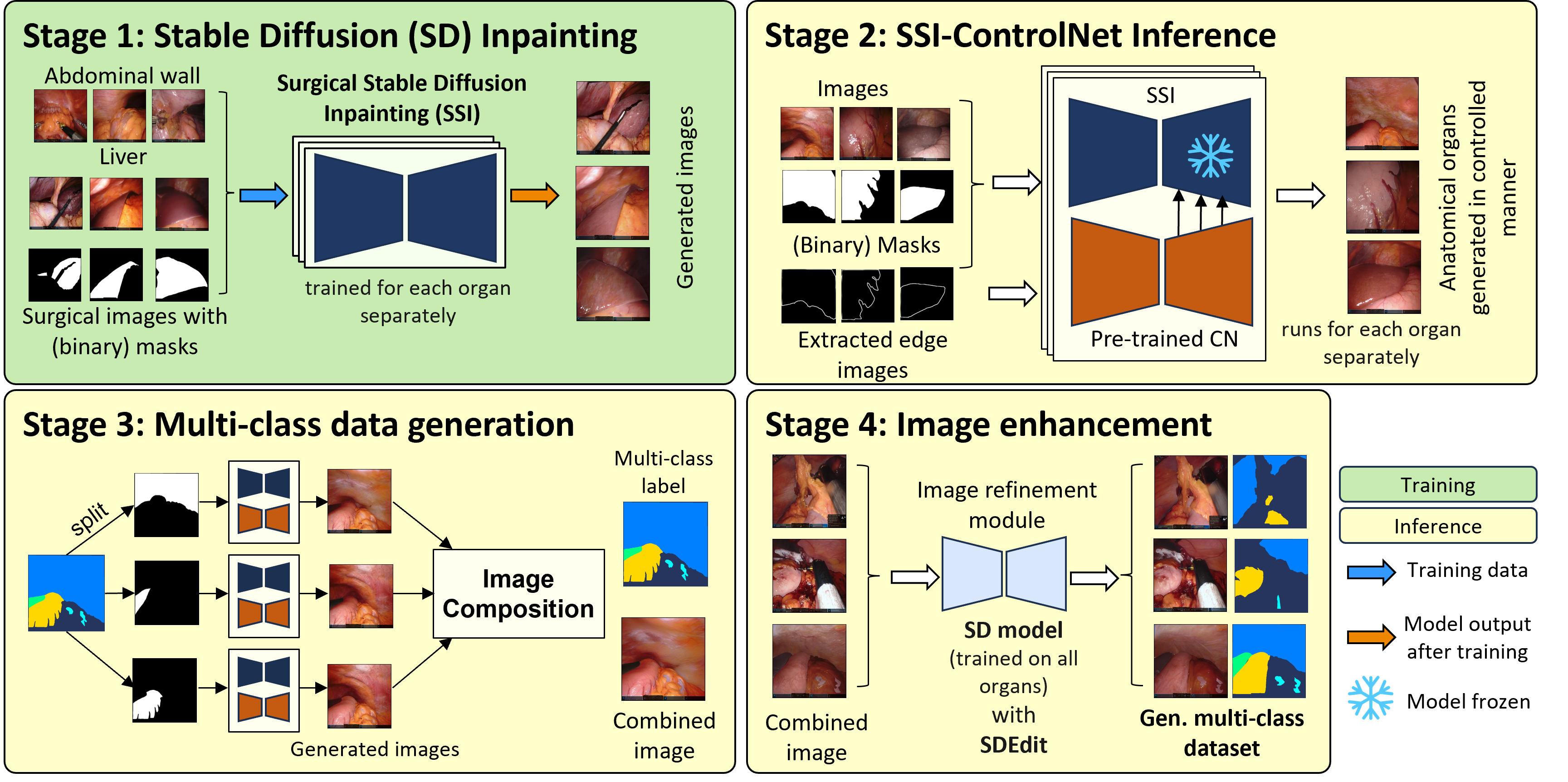}
\end{center}
   \caption{Overview of the diffusion approach to generate a multi-class dataset. Stage-$1$ involves training the SD inpainting model using the real images and masks for each organ separately. In stage-$2$, pre-trained ControlNet is plugged into the SSI model (SSI-CN) to precisely generate each anatomical structure using extracted edges from the segmentation mask. The image composition in stage-$3$ includes cutting out each organ from the generated image and combining them together to form the multi-class surgical dataset. Stage-$4$ (optional) includes an image refinement process using SDEdit~\cite{meng2021sdedit} to rectify inconsistencies during the composition operation and generate the multi-class images. We skip stage-$1$ for the simulated masks and start directly with the inference stages to generate the synthetic datasets.}
\label{fig:method}
\end{figure*}

To generate the multi-class datasets, we begin with real surgical images and segmentation masks to train the SD model using an inpainting objective. An inference pipeline containing this trained SD model and a pre-trained CN model is utilized to generate different organs. The generated organs are then fused into a multi-class image during the image composition stage. Finally, image refinement is performed using the SDEdit approach. An overview of the process is illustrated in~\cref{fig:method}.


\subsection{Stable Diffusion inpainting}\label{sec:sd-inpaint}
Our Stage-$1$ comprises training a diffusion inpainting model. Given an image $x$ with a mask $m$, the inpainting model is trained to generate an object in the masked region. Existing inpainting models randomly mask regions within an image. As a result, only partial objects or background areas can be generated in the masked region. Instead, we use the existing segmentation mask of the individual organs with the organ type as the corresponding text signal for training. In the forward process, the mask $m$ and its corresponding text prompt $P$ for image $x_0$ is used to obtain $\Tilde{x}_{t}$ with ~\cref{eq:1} as,
\begin{equation}
    \tilde{x}_t = x_t \odot m + x_0 \odot (1 - m),
\end{equation}
The model $\epsilon_\theta$ learns to predict the added noise $z_{t}$ with the objective
\begin{equation}
    \left.\mathcal{L}_{DM}=\mathbb{E}_{z \sim \mathcal{N}(0,I), t}\left[\| z-\epsilon_\theta\left(\Tilde{x}_t, t, P, m\right)\right) \|_2^2\right],
\end{equation}
We sample for $T$ steps starting from $x_T = \epsilon \odot m + x_0 \odot (1 - m)$ to obtain the inpainted result $x_0$. The inpainting objective focuses specifically on the masked region, which leads to the diffusion models learning the texture of each organ.
We call this model \textbf{S}urgical \textbf{S}table \textbf{I}npaint (SSI). We split the segmentation mask into $N$ individual organs and train the SSI model $N$ times separately. 
In this manner, we make our approach aware of each individual anatomy. As an additional flexibility this allows the introduction of new organs into a multi-class scene and only that diffusion model needs to be trained. Our approach adds minimal overhead in training compared to other methods based on GANs or diffusion models. Additionally, we fine-tune an SD model with only~\cref{eq:2} with all combined organs for the image refinement stage, which is explained below.

\subsection{Inference with ControlNet}
In Stage-$2$, anatomical structures are generated using CNs. Our preliminary results indicated that maintaining anatomical structures solely with segmentation masks and text prompts proved challenging. Hence, we opted for a simplified inference stage using pre-trained CN models. Training a CN model from scratch requires extensive computational resources and a large dataset. For instance, the CN model controlled by segmentation maps (CN-Seg) was trained with 164k images~\cite{zhang2023adding}. Hence, we circumvent this process by integrating a pre-trained CN model into the inpainting SSI (SSI-CN) model to control the shape and texture of the generated organs precisely. The number of classes for the pre-trained CN-Segmodel did not match our surgical datasets, so we opted for the pre-trained soft edge CN model, which uses extracted edge images from the segmentation masks as the conditioning signal. Given an input image and a mask, the new organ texture is inpainted only in the masked region leaving the background the same.

\subsection{Image composition}\label{smg}
In Stage-$3$, we generate the multi-class synthetic datasets. The different generated anatomical structures with SSI-CN model are cut out per organ from the generated image using the separate masks and combined to form the newly composed image. This results in an image comprising multiple classes with corresponding semantic labels. 

\subsection{Image enhancement stage}
\begin{figure}
\begin{center}
\includegraphics[width=0.4\textwidth]{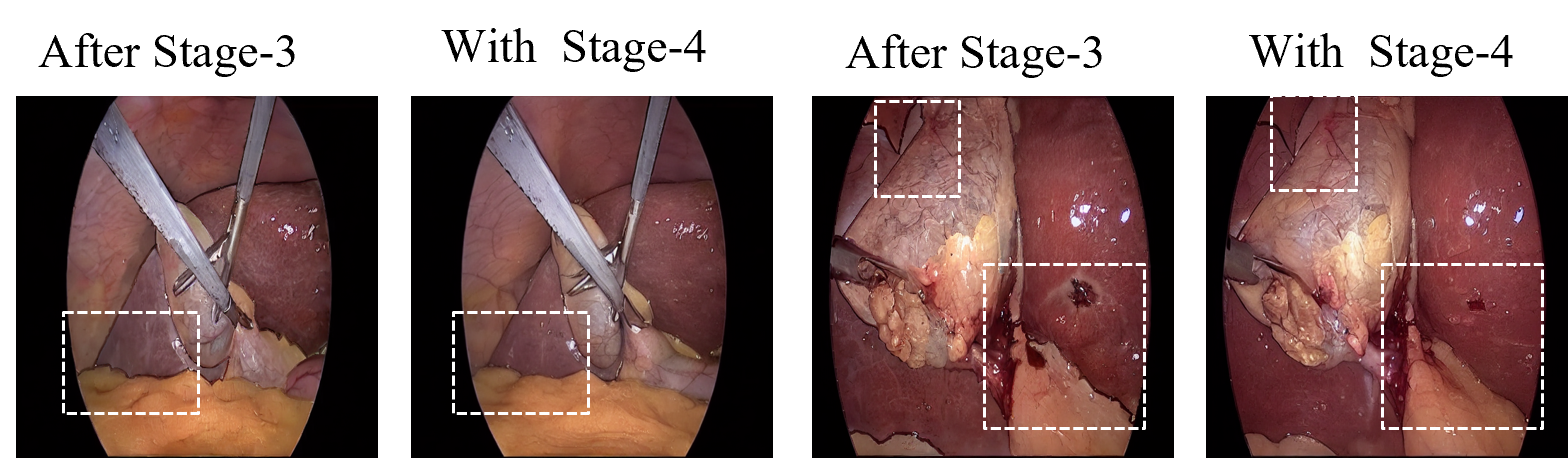}
\end{center}
   \caption{The generated images before and after Stage-$4$. White boxes show the inconsistent regions. The junction between two organs is smoothened, while the overall texture of the image is maintained.}
\label{fig:int}
\end{figure}

We noticed that the image composition operation introduced sharp edges between the organs and lighting artifacts, which is not present in real surgical images (see~\cref{fig:int}). Hence, in Stage-$4$, we perform an image enhancement step using SDEdit~\cite{meng2021sdedit}. We use the SD model trained with all organs combined with SDEdit to remove the inconsistencies introduced in the previous Stage-$3$. Low levels of noise has shown to improve texture components in images~\cite{si2024freeu} and hence this step can be optionally added to maintain the overall texture.

\section{Experiments \& Results}
In this section, we explain our experimental setup and the evaluation procedure for the generated synthetic images. We evaluate the generated datasets on image quality and their utility as training data for downstream segmentation.

\subsection{Data} 
For real surgical datasets, we used the CholecSeg8K~\cite{hong2020cholecseg8k} dataset, which is a labeled (multi-class) subset of Cholec80~\cite{endonet} with $5080$ training images and $2000$ test images, and the HeiSurf~\cite{bodenstedt2021heichole} dataset consisting of $330$ training and $110$ test images. Both these datasets involve the surgical removal of the gall bladder (Cholec.). Secondly, we utilize the DSAD dataset, which has binary (BN) segmentation masks of $1000$ images for each $11$ organs and a multi-class (MC) subset of $1400$ images containing six organs. We use the labels (six organs) common to both the datasets. It is to be noted that the MC-subset is smaller in number of images compared to the BN datasets. We used simulated masks from from Pfeiffer et al.~\cite{pfeiffer2019generating} and Rivoir et al.~\cite{Rivoir_2021_ICCV} for the Cholec. and DSAD datasets respectively.
\subsection{Task implementation} We design two distinct tasks to showcase our method's ability to work with multi- and binary-class datasets.

\textbf{Task T$1$: Multi $\rightarrow$ Multi} In this task, we use the multi-class segmentation masks from the CholecSeg8k and HeiSurf datasets. As explained in~\cref{sec:sd-inpaint}, the masks are split into binary classes for training and images are generated via fusion in Stage-$3$.

\textbf{Task T$2$: Binary $\rightarrow$ Multi} We re-iterate in this task on how our diffusion approach can be used to train only on real binary datasets and construct multi-class datasets. We use only the binary dataset (BN) from the DSAD dataset to train the diffusion models. For Stages $2$ and $3$, we use this dataset's multi-class (MC) segmentation masks as inputs and generate the multi-class synthetic datasets. 
For both tasks, we also use the simulated masks directly from Stage-$2$.

{\textbf{Implementation details}}
We use Stable-Diffusion inpainting v$1.5$ as the base diffusion model. The model was fine-tuned for $1500$ steps with a $1e^{-5}$ learning rate. The Soft Edge CN model was used during inference (Stage $2$) with a conditioning scale of $0.5$. For the details on the text prompts, kindly refer to the suppl. material. Our training setup for six organs plus takes $3.5$ hrs, whereas fine-tuning a pre-trained CN model already takes $3.2$ hrs on Nvidia RTX $A5000$ GPUs. An image is generated in $5.25$s using the inference pipeline(Stage-$2$ to Stage-$4$). For Stage-$4$, the SD model is trained for $4$ mins and the inference took $1.1$s, as only ten sampling steps were used for image refinement. We consider this to be a minimal overhead in comparison to annotating the surgical scenes.

\begin{table}
  \begin{center}
    {\small{
    \resizebox{\linewidth}{!}{
    \begin{tabular}{lcccc}
    \toprule
    Method & CFID ($\downarrow$) & KID ($\downarrow$) & CMMD($\downarrow$) & LPIPS($\downarrow$) \\
\hline
SPADE~\cite{park2019spade} & $391.85_{\pm6.53}$ & $0.39_{\pm0.04}$ & $1.94_{\pm0.13}$  & $0.60_{\pm0.03}$ \\
Pix2PixHD~\cite{wang2018high} & $371.81_{\pm8.81}$ & $0.50_{\pm0.02}$ & $2.47_{\pm0.06}$  & $0.72_{\pm0.02}$ \\
ControlNet~\cite{zhang2023adding} & $360.22_{\pm8.82}$ & $0.45_{\pm0.05}$  & $2.22_{\pm0.03}$ & $0.71_{\pm0.03}$ \\
T2i-Adapter~\cite{mou2023t2iadapter} & $358.50_{\pm8.57}$ & $0.56_{\pm0.03}$ & $4.17_{\pm0.15}$ & $0.76_{\pm0.03}$ \\ \hdashline
Ours-\emph{SS-Syn} & $\mathbf{337.16_{\pm3.23}}$ & $0.42_{\pm0.01}$ & $1.75_{\pm0.08}$ & $0.74_{\pm0.01}$ \\
Ours-\emph{Syn} & $348.04_{\pm9.10}$ & $\mathbf{0.39_{\pm0.03}}$ & $\mathbf{1.69_{\pm0.10}}$ & $\mathbf{0.58_{\pm0.03}}$ \\
    \bottomrule
\end{tabular}
}
}}
\end{center}
\caption{\textbf{Image quality comparison on CholecSeg8K dataset}. Our synthetic datasets show better quality than GAN or diffusion based approaches.}
\label{tab:cholec}
\end{table}

\subsection{Evaluation}
\textbf{Baselines} We compare our approach to popular semantic image synthesis GANs such as SPADE~\cite{park2019spade}, SPADE-vae~\cite{park2019spade} and Pix2PixHD~\cite{wang2018high}. 
For the diffusion approaches, we used the SD model trained on all organs as the base model and trained the ControlNet~\cite{zhang2023adding} and T2i-Adapter (T2i)~\cite{mou2023t2iadapter} from scratch conditioned on the multi-class labels from the real surgical datasets. We also fine-tuned the pre-trained Soft edge CN (ControlNet-SE) and Canny T2i (T2i-Adapter-CY). These models serve as powerful baselines for mask-conditioned image generation.\\

\textbf{Image quality} Firstly, we evaluate the quality and diversity of the generated images. We employ different metrics to compare the generated image quality. We use the popular metric CFID~\cite{parmar2022aliased} and KID~\cite{binkowski2018demystifying} 
to measure the realism of the images. The CMMD score~\cite{jayasumana2023rethinking} measures the image quality on better-extracted features from CLIP and is suitable for smaller datasets as it is unbiased, unlike CFID. Finally, we compute the LPIPS~\cite{zhang2018unreasonable} metric, highlighting the image's perceptual quality. \\

\begin{table}
  \begin{center}
    {\small{
    \resizebox{\linewidth}{!}{
\begin{tabular}{lccc}
\toprule
    Method 
    & Dice ($\uparrow$) & IOU ($\uparrow$) & HD($\downarrow$) \\
\hline
SPADE~\cite{park2019spade} & $0.58_{\pm0.01}$ & $0.46_{\pm0.02}$  & $119.30_{\pm1.75}$  \\
SPADE-vae~\cite{park2019spade} & $0.56_{\pm0.01}$ & $0.44_{\pm0.01}$  & $109.86_{\pm1.94}$ \\
Pix2Pix-HD~\cite{wang2018high} & $0.58_{\pm0.01}$ & $0.44_{\pm0.01}$ & $110.58_{\pm0.82}$ \\
ControlNet~\cite{zhang2023adding} & $0.57_{\pm0.01}$ & $0.44_{\pm0.01}$ & $115.59_{\pm6.80}$ \\
ControlNet-SE~\cite{zhang2023adding} & $0.61_{\pm0.01}$ & $0.48_{\pm0.02}$ & $107.54_{\pm3.49}$ \\
T2I-adapter~\cite{mou2023t2iadapter} & $0.59_{\pm0.02}$ & $0.46_{\pm0.01}$ & $117.55_{\pm1.63}$ \\
T2I-adapter-CY~\cite{mou2023t2iadapter} & $0.60_{\pm0.03}$ & $0.46_{\pm0.02}$ & $110.01_{\pm3.21}$ \\ \hdashline
Ours-\emph{SS-Syn} & $0.64_{\pm0.05}$ & $0.51_{\pm0.05}$ & $\mathbf{\mathbf{95.86_{\pm8.25}}}$ \\
Ours-\emph{Syn} & $\mathbf{0.68_{\pm0.01}}$ & $\mathbf{0.56_{\pm0.01}}$ & $95.93_{\pm6.89}$ \\
    \bottomrule
\end{tabular}
}
}}
\end{center}
\caption{\textbf{Segmentation comparison on the CholecSeg8K dataset} (T1:Multi$\rightarrow$Multi). The results of the model trained using our synthetic data outperforms all the baselines.}
\label{tab:cholec2}
\end{table}

\textbf{Downstream semantic segmentation}
We asses the utility of generated images by  performing two evaluations: $(1).$ To compare our approach to other image synthesis methods, we train a DV3+~\cite{chen2018encoder} model with the generated images from different image synthesis methods and fine-tune them on the real images. The model performance is evaluated on the test set of the real dataset. $(2).$ We train different state-of-the-art segmentation models such as Unet++~\cite{zhou2018unet},  UperNet~\cite{xiao2018unified} and Segformer~\cite{xie2021segformer} models. As baselines, we train each model using no augmentations, color augmentations, and a combination of color and spatial augmentations. We curated a list of different spatial and color augmentations from different segmentation competitions~\cite{cp1,cp2,nwoye2023cholectriplet2021} and prior works on medical image segmentation~\cite{goceri2023medical,kolbinger2023anatomy,jenke2024model,garcea2023data} which includes grid/elastic distortion, perspective change, RGB channel shift, ColorJitter, blur, hue, contrast \& brightness, maskdropout and train the models with them as baselines. For Task 2: Binary $\rightarrow$ Multi, we train implicit labeling method~\cite{jenke2024model} and fine-tune this model on the MC subset as a baseline. We compare the performance of these methods against the models that are trained on our datasets. \emph{Syn} denotes synthetic dataset using mask from the real surgical datasets and \emph{SS-Syn} uses masks from surgical simulations. Following suggestions from~\cite{maier2024metrics}, we chose Dice, IOU, and Hausdorff distance (HD) as the segmentation evaluation metrics and ignore the bg. as we inpaint only the masked organ region.
The readers can refer to the suppl. material for more details on the training process.

\subsection{Results}
\begin{figure}
\begin{center}
\includegraphics[width=0.4\textwidth]{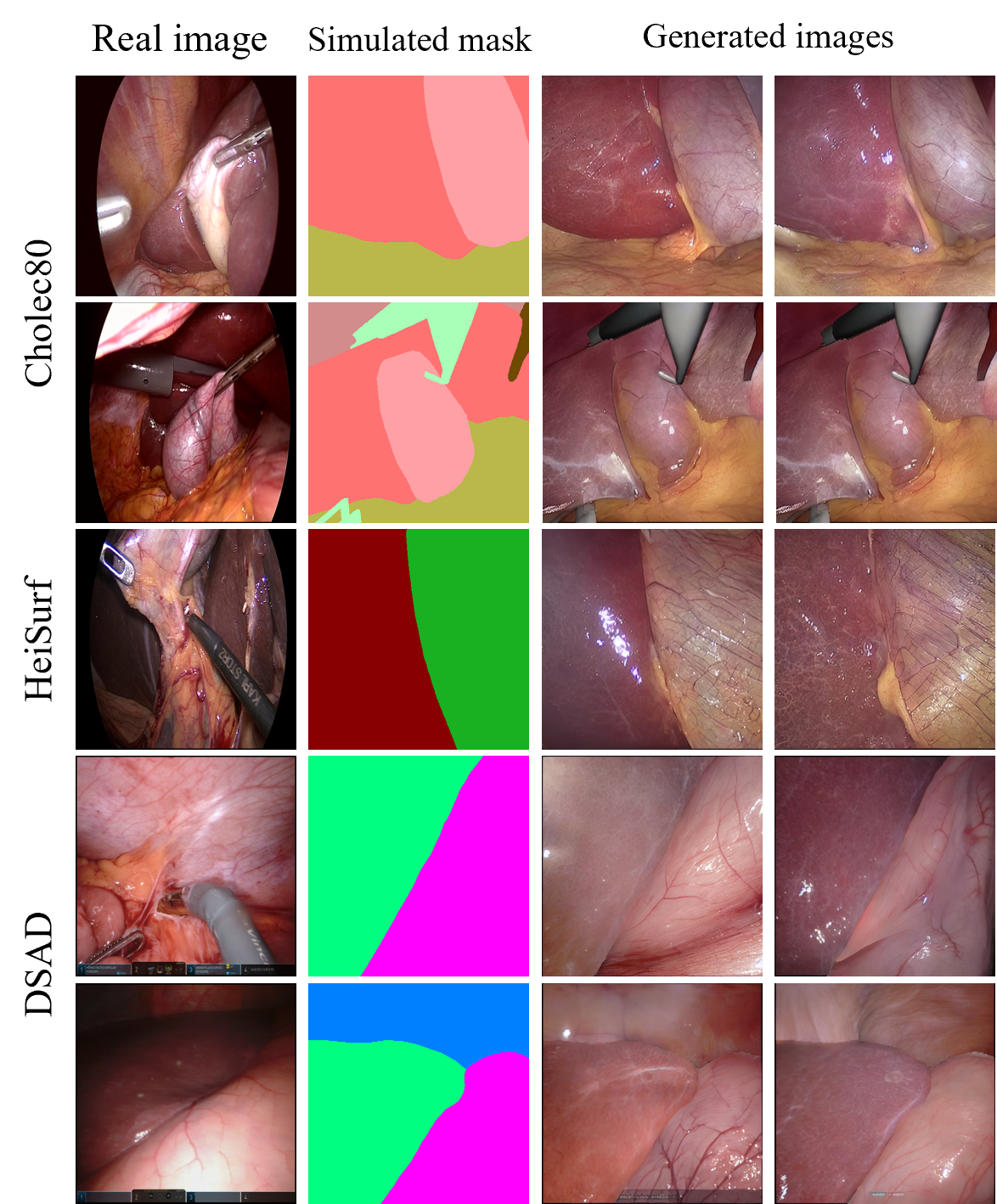}
\end{center}
   \caption{The generated images using simulated masks (SS). By using SS masks, we can generate surgical images other than the train datasets as the organ shapes differs with a similar organ texture to real datasets.}
\label{fig:syn_comp}
\end{figure}
\begin{table*}
  \begin{center}
  {\small{
    \resizebox{\linewidth}{!}{
\begin{tabular}{lccc|ccc|ccc}
\toprule
Training scheme &  \multicolumn{3}{c}{Unet$++$~\cite{zhou2018unet}} & \multicolumn{3}{c}{DV$3+$~\cite{chen2018encoder}} & \multicolumn{3}{c}{UperNet-Tiny~\cite{xiao2018unified}} \\
\hline
& Dice($\uparrow$) & IOU ($\uparrow$) & HD ($\downarrow$) & Dice ($\uparrow$) & IOU ($\uparrow$) & HD ($\downarrow$) & Dice ($\uparrow$) & IOU ($\uparrow$) & HD ($\downarrow$)\\
\cline{2-10}
Real with no-aug & $0.50_{\pm0.03}$ & $0.36_{\pm0.01}$ & $101.03_{\pm0.12}$ & $0.50_{\pm0.01}$ & $0.36_{\pm0.01}$ & $115.36_{\pm3.82}$ & $0.56_{\pm0.01}$ & $0.47_{\pm0.02}$ & $118.37_{\pm6.62}$\\
Real with color-aug & $0.52_{\pm0.01}$ & $0.38_{\pm0.02}$ & $\mathbf{98.95_{\pm2.05}}$ & $0.53_{\pm0.01}$ & $0.39_{\pm0.01}$ & $101.54_{\pm0.19}$ & $0.59_{\pm0.01}$ & $0.45_{\pm0.01}$ & $110.93_{\pm1.42}$\\
Real with color+spatial-aug & $0.61_{\pm0.05}$ & $0.49_{\pm0.04}$ & $109.09_{\pm0.52}$ & $0.58_{\pm0.01}$ & $0.45_{\pm0.01}$ & $108.14_{\pm1.07}$ & $0.61_{\pm0.04}$ & $0.50_{\pm0.05}$ & $108.63_{\pm1.51}$\\ \hdashline
Ours only \emph{Syn} & $0.53_{\pm0.03}$ & $0.40_{\pm0.01}$ & $110.65_{\pm1.31}$ & $0.53_{\pm0.01}$ & $0.41_{\pm0.02}$ & $108.66_{\pm1.18}$ & $0.56_{\pm0.01}$ & $0.44_{\pm0.01}$ & $109.41_{\pm2.09}$\\
Ours-\emph{SS-Syn} + Real & $\mathbf{0.67_{\pm0.01}}$ & $\mathbf{0.54_{\pm0.01}}$ & $107.10_{\pm0.49}$ & $0.64_{\pm0.05}$ & $0.51_{\pm0.05}$ & $\mathbf{95.86_{\pm8.25}}$ & $0.65_{\pm0.03}$ & $0.53_{\pm0.02}$ & $\mathbf{95.76_{\pm2.49}}$\\
Ours-\emph{Syn} + Real & $0.64_{\pm0.03}$ & $0.51_{\pm0.01}$ & $101.96_{\pm1.43}$ & $\mathbf{0.68_{\pm0.01}}$ & $\mathbf{0.56_{\pm0.01}}$ & $95.93_{\pm6.89}$ & $\mathbf{0.67_{\pm0.01}}$ & $\mathbf{0.54_{\pm0.01}}$ & $99.97_{\pm2.24}$\\

\bottomrule
\end{tabular}
}
}}
\end{center}
\caption{\textbf{Evaluation on CholecSeg8K dataset using different segmentation models} (T1:Multi$\rightarrow$Multi). A $10\%$ improvement was noticed in the segmentation scores with combined training \emph{Syn}+Real. The best scores are highlighted in bold.}
\label{tab:cholec3}
\end{table*}

\begin{figure*}
\begin{center}
\includegraphics[width=\textwidth]{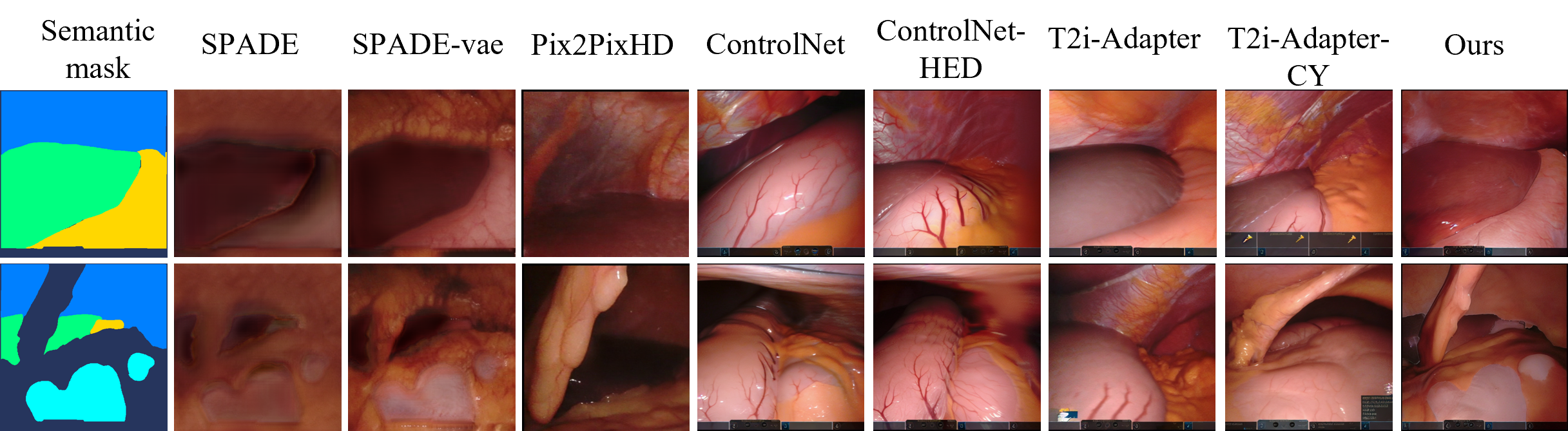}
\end{center}
   \caption{\textbf{Image quality comparison on the DSAD dataset.} The GAN methods (columns 2-4) fail to generate high quality images. The diffusion methods (columns 5-8) generate organs with realistic looking textures, however the spatial alignment to the semantic label is broken. Our method is able to maintain the shape and texture of different organs.}
\label{fig:dsad_comp}
\end{figure*}

\textbf{(Task $1$: Multi $\rightarrow$ Multi)}
$(1).$ \emph{CholecSeg8K}: The image quality evaluation results are indicated in~\cref{tab:cholec}. Overall, the results indicate that our approach generates high-quality and diverse surgical images.~\cref{fig:syn_comp} shows generated images from SS masks, further indicating that our approach can be used to generate organs with different shapes and textures that do not exist in the real datasets.~\cref{tab:cholec2} indicates the segmentation results using the generated images from different approaches. The ControlNet-SE shows a dice score of $0.61$ and IOU of $0.48$, which outperforms the GAN methods. Our \emph{Syn} dataset leads to $8\%$ improvement in dice and outperforms the ControlNet-SE model. The \emph{SS-Syn} dataset performs better than other baselines and falls slightly short of the \emph{Syn} dataset. 

The segmentation performance of different models using data augmentation is shown in~\cref{tab:cholec3}. Overall, adding color and spatial augmentations improved the performance across three models on the real surgical datasets. The results show that only using the \emph{Syn} dataset already matches the scores of the real images. We consistently see performance improvement across all the segmentation models when combined \emph{Syn}+Real training is done. The \emph{SS-Syn}+Real dataset scores are similar to those of the \emph{Syn} datasets. We hypothesize that the remaining performance difference can be attributed to the organ shape between the domains.

\begin{table}
  \begin{center}
    {\small{
    \resizebox{\linewidth}{!}{
\begin{tabular}{lcccc}
\toprule
    Method & CFID ($\downarrow$) & KID ($\downarrow$) & CMMD($\downarrow$) & LPIPS($\downarrow$) \\
\hline
SPADE~\cite{park2019spade} & $382.55_{\pm2.32}$ & $0.35_{\pm0.05}$ & $2.08_{\pm0.04}$  & $0.70_{\pm0.04}$ \\
Pix2PixHD~\cite{wang2018high} & $\mathbf{347.75_{\pm4.12}}$ & $0.39_{\pm0.04}$ & $3.81_{\pm0.10}$  & $0.82_{\pm0.04}$ \\
ControlNet-SE~\cite{zhang2023adding} & $380.68_{\pm3.41}$ & $0.32_{\pm0.05}$  & $2.09_{\pm0.03}$ & $0.85_{\pm0.03}$ \\
T2i-Adapter-CY~\cite{mou2023t2iadapter} & $409.16_{\pm3.67}$ & $0.34_{\pm0.05}$ & $1.07_{\pm0.05}$ & $0.78_{\pm0.03}$ \\ \hdashline
Ours-\emph{SS-Syn} & $351.09_{\pm2.70}$ & $0.39_{\pm0.05}$ & $0.69_{\pm0.03}$ & $0.69_{\pm0.04}$ \\
Ours-\emph{Syn} & $369.62_{\pm8.15}$ & $\mathbf{0.31_{\pm0.01}}$ & $\mathbf{0.57_{\pm0.02}}$ & $\mathbf{0.68_{\pm0.01}}$ \\
    \bottomrule
\end{tabular}
}
}}
\end{center}
\caption{\textbf{Image quality comparison on HeiSurf dataset}. The images generated from our method shows better image quality than other methods.}
\label{tab:hs}
\end{table}

\begin{table}
\resizebox{\linewidth}{!}{
\begin{tabular}{lccc|ccc}
\toprule
Training scheme &  \multicolumn{3}{c}{Unet$++$~\cite{zhou2018unet}} & \multicolumn{3}{c}{DV$3+$~\cite{chen2018encoder}} \\
\hline
& Dice($\uparrow$) & IOU ($\uparrow$) & HD ($\downarrow$) & Dice ($\uparrow$) & IOU ($\uparrow$) & HD ($\downarrow$) \\
\cline{2-7}
Real with no-aug & $0.40_{\pm0.02}$ & $0.29_{\pm0.01}$ & $\mathbf{148.36_{\pm2.56}}$ & $0.30_{\pm0.05}$ & $0.20_{\pm0.07}$ & $239.57_{\pm9.69}$ \\
Real with color-aug & $0.42_{\pm0.01}$ & $0.30_{\pm0.02}$ & $165.76_{\pm2.02}$ & $0.42_{\pm0.01}$ & $0.30_{\pm0.02}$ & $180.41_{\pm3.61}$ \\
Real with color+spatial-aug & $0.45_{\pm0.01}$ & $0.32_{\pm0.01}$ & $247.21_{\pm8.15}$ & $0.40_{\pm0.02}$ & $0.31_{\pm0.01}$ & $206.34_{\pm3.04}$ \\ \hdashline
Ours only \emph{Syn} & $0.42_{\pm0.01}$ & $0.30_{\pm0.01}$ & $201.20_{\pm9.01}$ & $0.35_{\pm0.02}$ & $0.24_{\pm0.01}$ & $205.34_{\pm3.62}$ \\
Ours-\emph{SS-Syn} + Real & $\mathbf{0.55_{\pm0.01}}$ & $0.36_{\pm0.02}$ & $211.81_{\pm2.47}$ & $0.47_{\pm0.01}$ & $0.33_{\pm0.01}$ & $170.63_{\pm2.19}$ \\
Ours-\emph{Syn} + Real & $0.53_{\pm0.01}$ & $\mathbf{0.40_{\pm0.01}}$ & $207.65_{\pm1.70}$ & $\mathbf{0.49_{\pm0.01}}$ & $\mathbf{0.36_{\pm0.01}}$ & $\mathbf{165.42_{\pm3.04}}$ \\

\bottomrule
\end{tabular}
}
\caption{\textbf{Comparison of data augmentations on HeiSurf dataset} (T$1$:Multi$\rightarrow$Multi). Using the generated images leads to improved performance across the two models.}
\label{tab:hs2}
\end{table}

\begin{table*}
\parbox{.40\linewidth}{

\resizebox{\linewidth}{!}{
\begin{tabular}{lccc}
\toprule
    Method & Dice ($\uparrow$) & IOU ($\uparrow$) & HD($\downarrow$) \\
\hline
SPADE~\cite{park2019spade} & $0.75_{\pm0.01}$ & $0.67_{\pm0.04}$ & $93.63_{\pm0.33}$ \\
SPADE-vae~\cite{park2019spade} & $0.79_{\pm0.02}$ & $0.69_{\pm0.01}$ & $\mathbf{83.39_{\pm4.50}}$ \\
Pix2Pix-HD~\cite{wang2018high} & $0.76_{\pm0.02}$ & $0.66_{\pm0.01}$ & $103.66_{\pm1.02}$ \\
ControlNet~\cite{zhang2023adding} & $0.76_{\pm0.03}$ & $0.67_{\pm0.01}$ & $101.67_{\pm3.71}$ \\
ControlNet-SE~\cite{zhang2023adding} & $0.78_{\pm0.02}$ & $0.68_{\pm0.01}$ & $95.53_{\pm3.08}$ \\
T2I-adapter~\cite{mou2023t2iadapter} & $0.78_{\pm0.03}$ & $0.68_{\pm0.01}$ & $89.15_{\pm3.89}$ \\
T2I-adapter-CY~\cite{mou2023t2iadapter} & $0.76_{\pm0.01}$ & $0.67_{\pm0.03}$ & $89.28_{\pm2.36}$ \\ \hdashline
Ours-\emph{SS-Syn} & $0.82_{\pm0.01}$ & $0.73_{\pm0.04}$ & $86.32_{\pm4.15}$ \\
Ours-\emph{Syn} & $\mathbf{0.83_{\pm0.01}}$ & $\mathbf{0.74_{\pm0.01}}$ & $92.50_{\pm6.05}$ \\
    \bottomrule
\end{tabular}
}
\caption{\textbf{Segmentation eval. on DSAD dataset.} Our synthetic datasets show superior performance to other baselines.}
\label{tab:dsad_comp}
}
\hfill
\parbox{.55\linewidth}{
\resizebox{\linewidth}{!}{
\begin{tabular}{lccc|ccc}
\hline
Training scheme & \multicolumn{3}{c}{DV$3+$~\cite{chen2018encoder}} & \multicolumn{3}{c}{Segformer~\cite{xie2021segformer}} \\
\midrule
& Dice($\uparrow$) & IOU ($\uparrow$) & HD ($\downarrow$) & Dice ($\uparrow$) & IOU ($\uparrow$) & HD ($\downarrow$)\\
\cline{2-7}
Real (MC) with no-aug & $0.78_{\pm0.01}$ & $0.68_{\pm0.03}$ & $218.80_{\pm7.73}$ & $0.81_{\pm0.01}$ & $0.74_{\pm0.02}$ & $73.80_{\pm2.52}$\\
Real (MC) with color-aug & $0.78_{\pm0.03}$ & $0.70_{\pm0.02}$ & $206.07_{\pm5.26}$ & $0.83_{\pm0.01}$ & $0.75_{\pm0.01}$ & $75.79_{\pm2.01}$\\
Real (MC) with color+spatial-aug & $0.79_{\pm0.02}$ & $0.70_{\pm0.02}$ & $196.24_{\pm7.24}$ & $0.84_{\pm0.02}$ & $0.76_{\pm0.01}$ & $85.58_{\pm7.49}$\\ \hdashline
Implicit label~\cite{jenke2024model} on Real (BN) & $0.22_{\pm0.02}$ & $0.13_{\pm0.01}$ & $296.45_{\pm5.96}$ & $0.22_{\pm0.01}$ & $0.15_{\pm0.01}$ & $324.90_{\pm8.85}$\\
Implicit label on Real (BN) + Real (MC) & $0.80_{\pm0.01}$ & $0.70_{\pm0.01}$ & $83.56_{\pm2.01}$ & $0.82_{\pm0.01}$ & $0.74_{\pm0.01}$ & $74.31_{\pm6.71}$\\ \hdashline
Ours only \emph{Syn} & $0.60_{\pm0.03}$ & $0.51_{\pm0.01}$ & $95.19_{\pm1.11}$ & $0.62_{\pm0.01}$ & $0.51_{\pm0.02}$ & $116.14_{\pm1.08}$\\
Ours \emph{Syn} + Implicit label on Real (BN) + Real (MC) & $0.81_{\pm0.01}$ & $0.70_{\pm0.02}$ & $\mathbf{81.32_{\pm4.59}}$ & $0.84_{\pm0.02}$ & $0.76_{\pm0.02}$ & $\mathbf{69.52_{\pm3.35}}$\\
Ours \emph{SS-Syn} + Real (MC) & $0.82_{\pm0.01}$ & $0.73_{\pm0.04}$ & $86.32_{\pm4.15}$ & $0.84_{\pm0.01}$ & $0.75_{\pm0.02}$ & $82.54_{\pm5.20}$\\
Ours \emph{Syn} + Real (MC) & $\mathbf{0.83_{\pm0.01}}$ & $\mathbf{0.74_{\pm0.01}}$ & $92.50_{\pm6.05}$ & $\mathbf{0.86_{\pm0.01}}$ & $\mathbf{0.78_{\pm0.02}}$ & $79.90_{\pm1.04}$\\

\bottomrule
\end{tabular}
}
\caption{\textbf{Comparison of augmentations on DSAD dataset} (T$2$:Binary$\rightarrow$Multi). BN denotes binary dataset and MC multi-class subset. The combined \emph{Syn}+Real on (MC) training method shows the best performance across the two models.}
\label{tab:dsad_final}
}
\end{table*}

$(2).$ HeiSurf: The results on image quality in~\cref{tab:hs} indicates that our approach is better in maintaining realism of the images. 
For such smaller datasets, other image synthesis methods suffer to generate images suitable for the application. Our \emph{Syn} datasets leads to a $10\%$ difference in scores compared to other models (in \emph{suppl}). Furthermore as evidenced from~\cref{tab:hs2}, for the Unet++ architecture, the combined training \emph{Syn}+Real shows a $8\%$ improvement in both dice and IOU compared to the data augmentations on the real images. For the DV3+ model, dice score improved by $9\%$ with a drastic improvement in HD scores when using the \emph{Syn} dataset. These results further show that our approach is effective at capturing the texture of different organs, thereby allowing the generation of surgical datasets. Additonal qualitative results are in suppl. material.

\textbf{(Task 2: Binary $\rightarrow$ Multi)}\label{res:2}
The qualitative results are presented in~\cref{fig:dsad_comp}. Our method precisely generates the organs according to the semantic mask. In contrast, the GAN-based method fails to maintain image quality, and the diffusion approaches fall short in maintaining spatial alignment. These results highlight the importance of the image composition stage, which aids in preserving the organ structures while the diffusion process effectively generates their textures. The segmentation scores shown in~\cref{tab:dsad_comp} indicate that our method outperforms the baselines, achieving an improvement of more than $8\%$ in scores. Additionally, in~\cref{tab:dsad_final}, the results demonstrate that combining the generated synthetic data leads to a $5\%$ improvement in dice and IOU for the DV3+ model, with a notable boost in HD scores observed for the Segformer model. Combining the generated datasets with the implicit labeling method showed smaller improvements.

\textbf{Ablation study}
The results of the ablation study is shown in~\cref{tab:abl}. In~\cref{tab:abl} Config A, we removed the pre-trained CN from Stage-$2$ and the image enhancement stage (Stage-$4$) on the DSAD datasets. A decrease in segmentation metrics is seen in this case. Similarly, we added Stage-$4$ for Config B and saw minor improvements in the scores. Config C, our approach highlights the need for a combination of both the stages shown by higher dice and HD scores.
\begin{table}
  \begin{center}
    {\small{
    \resizebox{\linewidth}{!}{
\begin{tabular}{lcccccc}
\toprule
    Method & Stage-$2$ CN & Stage-$4$  & Dice ($\uparrow$) & IOU($\uparrow$) & HD($\downarrow$) \\
\hline
Config A & $\times$ & $\times$ & $0.52_{\pm0.01}$ & $0.41_{\pm0.01}$ & $110.55_{\pm4.58}$ \\ 
Config B & $\times$ & \checkmark & $0.55_{\pm0.02}$ & $0.42_{\pm0.01}$ & $102.10_{\pm2.42}$ \\ 
Ours & \checkmark & \checkmark & $0.60_{\pm0.03}$ & $0.51_{\pm0.01}$ & $95.19_{\pm1.11}$\\ 
\end{tabular}
}
}}
\end{center}
\caption{\textbf{Ablation study on the DSAD dataset.} Removing Stage-$4$ or the CN from Stage-$2$ leads to drop in performance.}
\label{tab:abl}
\end{table}

\section{Discussion and Limitations}
In this study, we present a diffusion approach for generating multi-class surgical datasets. Our findings show that guiding the generation process through inpainting and edge-conditioning organ-specific models preserves both the texture and shape of the organs. Moreover, our approach allows the creation of multi-class datasets using only binary real datasets and multi-class simulated masks. The evaluation results confirm that our synthetic datasets are high quality and valuable as downstream training datasets. As our generated images contain realistic textures they can be effectively utilized as pre-training datasets for other downstream surgical tasks like surgical target prediction or detection (Tab.5 in \emph{suppl.}).

\textbf{Limitations}. Our approach produces realistic images, but it has limitations. We train diffusion models on each organ separately. Implementing spatial conditioning of subjects like bounded attention~\cite{dahary2024yourself} could be a promising solution for multi-organ composition. Due to organ compositionality we fall behind on maintaining the depth of the generated images. As a promising alternative pseudo depth maps could be integrated into the inference pipeline. Furthermore, to improve the time taken to generate the datasets, advanced noise schedulers~\cite{luo2023latent,lu2022dpm} using few sampling steps can be explored. 

\textbf{Acknowledgements}. This work is partly supported by BMBF (Federal Ministry of Education and Research) in DAAD project 57616814 (\href{https://secai.org/}{SECAI, School of Embedded Composite AI}). This work is partly funded by the German Research Foundation (DFG, Deutsche Forschungsgemeinschaft) as part of Germany’s Excellence Strategy – EXC 2050/1 –Project ID 390696704 – Cluster of Excellence “Centre for Tactile Internet with Human-in-the-Loop” (CeTI) of Technische Universit¨at Dresden. DK. thanks Isabel Funke for the insightful discussions.
{\small
\bibliographystyle{ieee_fullname}
\bibliography{ref}
}

\begin{center}
    \textbf{\large Supplementary material}
\end{center}
\setcounter{section}{0}
\renewcommand{\thesection}{\Alph{section}}
\section{Diffusion training details}
We use the Stable Diffusion v$1.5$ inpainiting model as our baseline diffusion model. The training parameters for the CholecSeg8K dataset are shown in~\cref{tab:cholec}. The dataset consists of $5080$ images for training, $2000$ images for testing and $1000$ images as validation set. We resized all the images to $512$x$512$. Depending upon our initial experiments, we noticed for the abdominal wall that the generated images either suffered from creating the correct texture or having semantic leakage i.e., texture of other organs being replaced. One reason could be the variance in the lighting conditions. To account for this, be opted for the $v$-prediction loss from~\cite{salimans2022progressive}. This has shown to improve image quality in low lighting. For the HeiSurf dataset, we used the parameters as the CholecSeg8K dataset. All the diffusion models were trained $1500$ steps. The generated images were evaluated at every 500 steps using different metrics and visual examination. The pre-trained soft edge ControlNet was used to control the anatomy shape during the inference process. We sampled images from the inpainted model with pre-trained CN with DDIM~\cite{song2022denoising} scheduler using $30$ steps. A similar process was used for the DSAD dataset where a guidance scale of $5.5$ was used for all the organs. For this dataset, we did not use the v-prediction method for training any specific organ. To reduce the overhead in the inference pipeline, we used the fast MultiStep~\cite{zhao2023uni} sampler in Stage-$4$ with SDEdit for image enchancement. We opted for $5-8$ steps for inconsistency removal process. Since the HeiSurf dataset had smaller image of images, we trained only the ControlNet-SE and T2i-Adapter-CY models on this dataset.

\textbf{Text prompts} For our approach, we used simple text prompts like \emph{an image of abdominal wall in cholec} for the abdominal wall images in the CholecSeg8K dataset. Similarly, for the other organs and datasets we exchanged the organ name and the dataset name accordingly. For the SD model used in Stage-$4$, the text prompt was chosen as \emph{an image in cholec} for an image in CholecSeg8K. 

To train the ControlNet and T2i-Adapters pre-trained SD model is needed. We experimented with different text prompts. Intially, we used the same prompts from our method like \emph{an image of cholec surgery} to train the SD model. This model was then used to train the ControlNet and T2i models. For their training, we again used the same prompts as the SD model. We noticed that the generated images lacked quality and did not correspond well to the conditioning masks. Hence, we used the segmentation masks to extract the classes present and constructed the prompt like \emph{an image of cholec surgery with abdominal wall, liver and gall bladder with a hook}. We train the the SD model with such prompts and use similar prompts for training the ControlNet and T2i models. We found the best results with such expressive prompts rather than just mentioning a prompt like \emph{an image of cholec surgery}. We hypothesize such prompts are necessary to make the model explicitly understand the different organs present in the scene. It is to noted that extra effort in constructing such prompts were necessary to train the baseline models in comparison our model which works on simpler text prompts. As we had limits on our training infrastructure, we did not train the text encoder of these models. As a future work we intend to train the train encoder along with the diffusion models to scope their performance on image quality.

\begin{table}
  \begin{center}
    {\small{
    \resizebox{\linewidth}{!}{
    \begin{tabular}{lcccc}
    \toprule
    Organ & Pred-type & Gd. scale \\
\hline
Abdominal wall & v-prediction & $0.6$ \\
Fat & $\epsilon$-prediction & $5.0$ \\
Liver & $\epsilon$-prediction & $6.0$ \\
Gall bladder & $\epsilon$-prediction & $5.5$ \\
Ligament & $\epsilon$-prediction & $5.0$ \\
    \bottomrule
\end{tabular}
}
}}
\end{center}
\caption{The parameters used for training and sampling from the CholecSeg8K dataset.}
\label{tab:cholec}
\end{table}

\begin{table}
  \begin{center}
  {\small{
    \resizebox{\linewidth}{!}{
\begin{tabular}{lccc}
\hline
Training scheme &  \multicolumn{3}{c}{Unet$++$~\cite{zhou2018unet}} \\
\hline
& Dice($\uparrow$) & IOU ($\uparrow$) & HD ($\downarrow$) \\
\cline{2-4}
No-aug & $0.74_{\pm0.03}$ & $0.65_{\pm0.01}$ & $126.08_{\pm3.02}$ \\
Color-aug & $0.76_{\pm0.01}$ & $0.66_{\pm0.02}$ & $118.98_{\pm1.32}$ \\
Color+spatial-aug & $0.79_{\pm0.01}$ & $0.69_{\pm0.01}$ & $88.86_{\pm9.93}$ \\ \hdashline
Implicit label~\cite{jenke2024model}& $0.22_{\pm0.05}$ & $0.11_{\pm0.01}$ & $347.33_{\pm9.12}$ \\
Implicit label + Real & $0.77_{\pm0.04}$ & $0.67_{\pm0.03}$ & $97.44_{\pm2.75}$ \\ \hdashline
Only \emph{Syn} & $0.44_{\pm0.03}$ & $0.34_{\pm0.01}$ & $132.63_{\pm4.16}$\\
\emph{Syn} + Implicit label + Real & $0.75_{\pm0.02}$ & $0.65_{\pm0.03}$ & $93.82_{\pm1.87}$ \\
\emph{SS-Syn} + Real & $0.80_{\pm0.03}$ & $0.70_{\pm0.02}$ & $102.01_{\pm3.34}$ \\
\emph{Syn} + Real & $\mathbf{0.82_{\pm0.01}}$ & $\mathbf{0.72_{\pm0.01}}$ & $\mathbf{85.27_{\pm1.04}}$ \\

\bottomrule
\end{tabular}
}
}}
\end{center}
\caption{The segmentation scores on the DSAD dataset. The best scores are indicated in bold.}
\label{tab:dsad_final}
\end{table}

\section{Segmentation training details}
To train the baselines on different augmentation schemes, we collected and experimented with multiple methods. We curated a set of color and spatial augmentations based on prior works that focussed on medical (surgical) domain~\cite{jenke2024model,kolbinger2023anatomy,goceri2023medical}. We used the following augmentations:grid/elastic distortion, perspective change, RGB channel shift, ColorJitter, blur, hue, contrast \& brightness, maskdropout. We tuned the hyperparameters included in each of these methods to attain the best scores. Similarly, for spatial transformations we used perspective change, grid distortion, rotation, random flipping. For the combined (color+spatial) augmentations, we chose the best combination via experimentation with different combinations of methods mentioned before. To find the best combinations of methods and hyperparameters,  we conducted experiments on each dataset separately.

\section{Additional results}
The segmentation results on the DSAD dataset with Unet++ architecture is shown in~\cref{tab:dsad_final}. Similarly, for the CholecSeg8K dataset, we also trained the UperNet-small model. The results are shown in~\cref{tab:ch_final}. The seg. scores from different image synthesis models for the HeiSurf dataset is shown in~\cref{tab:hs-seg}. 

\underline{\textbf{Auxillary surgical task}}. We used the generated datasets to train models for another surgical task: \emph{surgical target prediction.} We used the CholecT$50$~\cite{nwoye2022rendezvous} as it forms a part of CholecSeg8K and DSAD datasets to show the capability of our \emph{Syn} dataset in multi-class and multi-label classification tasks. The training and test splits were maintained throughout to avoid any data leakage.~\cref{tab:targ} shows that our \emph{Syn} datasets proves useful beyond segmentation tasks. 

The results in~\cref{tab:sched} shows using UniPc~\cite{zhao2024unipc} scheduler with $20$ sampling steps. This leads to the inference time of $4.07$s in comparison to $5.25$s. We also notice that the downstream performance of the generated images matches that of the DDIM scheduler.

\begin{table}
  \begin{center}
  {\small{
    \resizebox{\linewidth}{!}{
\begin{tabular}{lccc}
\hline
Training scheme &  \multicolumn{3}{c}{UperNet-small} \\
\hline
& Dice($\uparrow$) & IOU ($\uparrow$) & HD ($\downarrow$) \\
\cline{2-4}
No-aug & $0.55_{\pm0.01}$ & $0.47_{\pm0.02}$ & $118.37_{\pm6.62}$ \\
Color-aug & $0.57_{\pm0.01}$ & $0.45_{\pm0.04}$ & $115.80_{\pm1.38}$ \\
Color+spatial-aug & $0.62_{\pm0.02}$ & $0.51_{\pm0.01}$ & $108.63_{\pm1.51}$ \\ \hdashline
Only \emph{Syn} & $0.56_{\pm0.01}$ & $0.45_{\pm0.01}$ & $111.71_{\pm0.62}$\\
\emph{SS-Syn} + Real & $\mathbf{0.69_{\pm0.01}}$ & $\mathbf{0.53_{\pm0.02}}$ & $\mathbf{95.76_{\pm2.49}}$ \\
\emph{Syn} + Real & $0.67_{\pm0.01}$ & $\mathbf{0.53_{\pm0.01}}$ & $105.90_{\pm5.28}$ \\

\bottomrule
\end{tabular}
}
}}
\end{center}
\caption{The segmentation scores on the CholecSeg8K dataset. The best scores are indicated in bold.}
\label{tab:ch_final}
\end{table}

\begin{table}
\resizebox{\linewidth}{!}{
\begin{tabular}{lccc}
\toprule
    Method & Dice ($\uparrow$) & IOU ($\uparrow$) & HD($\downarrow$) \\
\hline
SPADE~\cite{park2019spade} & $0.39\pm_{0.01}$ & $0.27_{\pm0.01}$ & $252.70_{\pm7.17}$ \\
SPADE-vae~\cite{park2019spade}  & $0.39_{\pm0.01}$ & $0.28_{\pm0.02}$ & $234.89_{\pm6.15}$ \\
Pix2Pix-HD~\cite{wang2018high} & $0.39_{\pm0.03}$ & $0.27_{\pm0.02}$ & $236.35_{\pm8.73}$ \\
ControlNet-SE~\cite{zhang2023adding} & $0.40_{\pm0.02}$ & $0.27_{\pm0.03}$ & $224.28_{\pm5.02}$ \\
T2I-adapter-CY~\cite{mou2023t2iadapter} & $0.38_{\pm0.01}$ & $0.26_{\pm0.01}$ & $234.97_{\pm8.32}$ \\ \hdashline
Ours-\emph{SS-Syn} & $0.47_{\pm0.01}$ & $0.33_{\pm0.01}$ & $170.63_{\pm2.19}$ \\
Ours-\emph{Syn} & $\mathbf{0.49_{\pm0.01}}$ & $\mathbf{0.36_{\pm0.01}}$ & $\mathbf{165.42_{\pm3.04}}$ \\
    \bottomrule
\end{tabular}
}
\caption{\textbf{Segmentation eval. on HeiSurf dataset.} Our synthetic datasets outperforms other models.}
\label{tab:hs-seg}
\end{table}

\begin{figure*}[ht]
\begin{center}
\includegraphics[width=\textwidth]{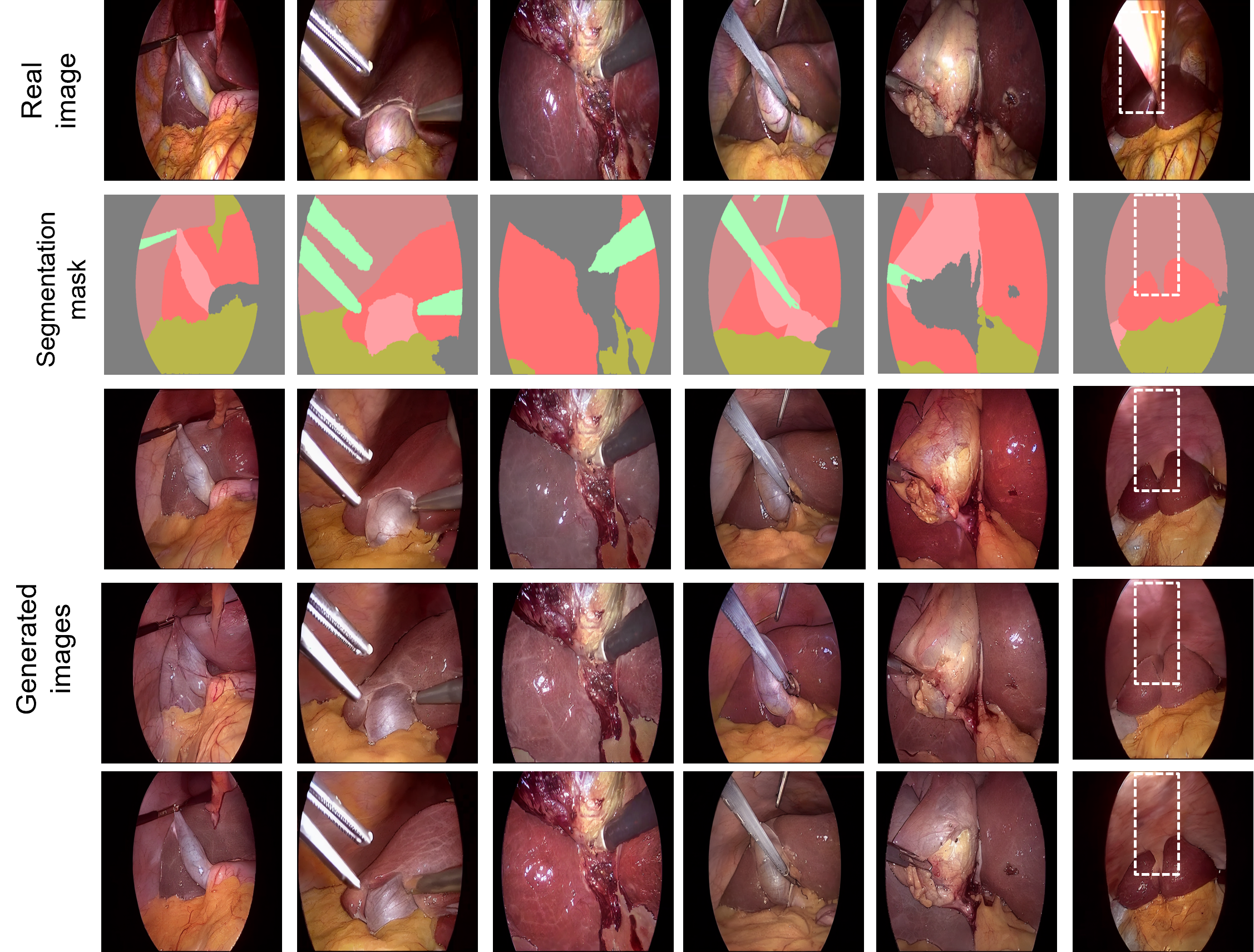}
\end{center}
   \caption{The generated images from the CholecSeg8K dataset. The different organs generated by our method followed by fusion creates images looking similar to real images. The diversity in the texture of the generated organs are quite visible in these images(zoom into the images to see the texture difference). Column $3,4,5$ clearly shows the difference in the liver and gall bladder textures. In the $6^{th}$ column we see that the generated images differ (indicated in white box) to the real image. This is because the ligament (bright yellow organ) in the real image was not labeled due to the camera angle and the light source and rather a common label of abdominal wall was indicated. Since our approach uses the segmentation masks for generating the organs, the ligament is not generated in the images, which does not affect downstream performance as the generated image still corresponds to the label. We see this an avenue for future work rather than a limitation. Using surgical simulations, either ligament or new organs can be generated using our diffusion approach, wherein only one model needs to be trained on that specific organ.}
\label{fig:ch}
\end{figure*}

\begin{figure*}[ht]
\begin{center}
\includegraphics[width=\textwidth]{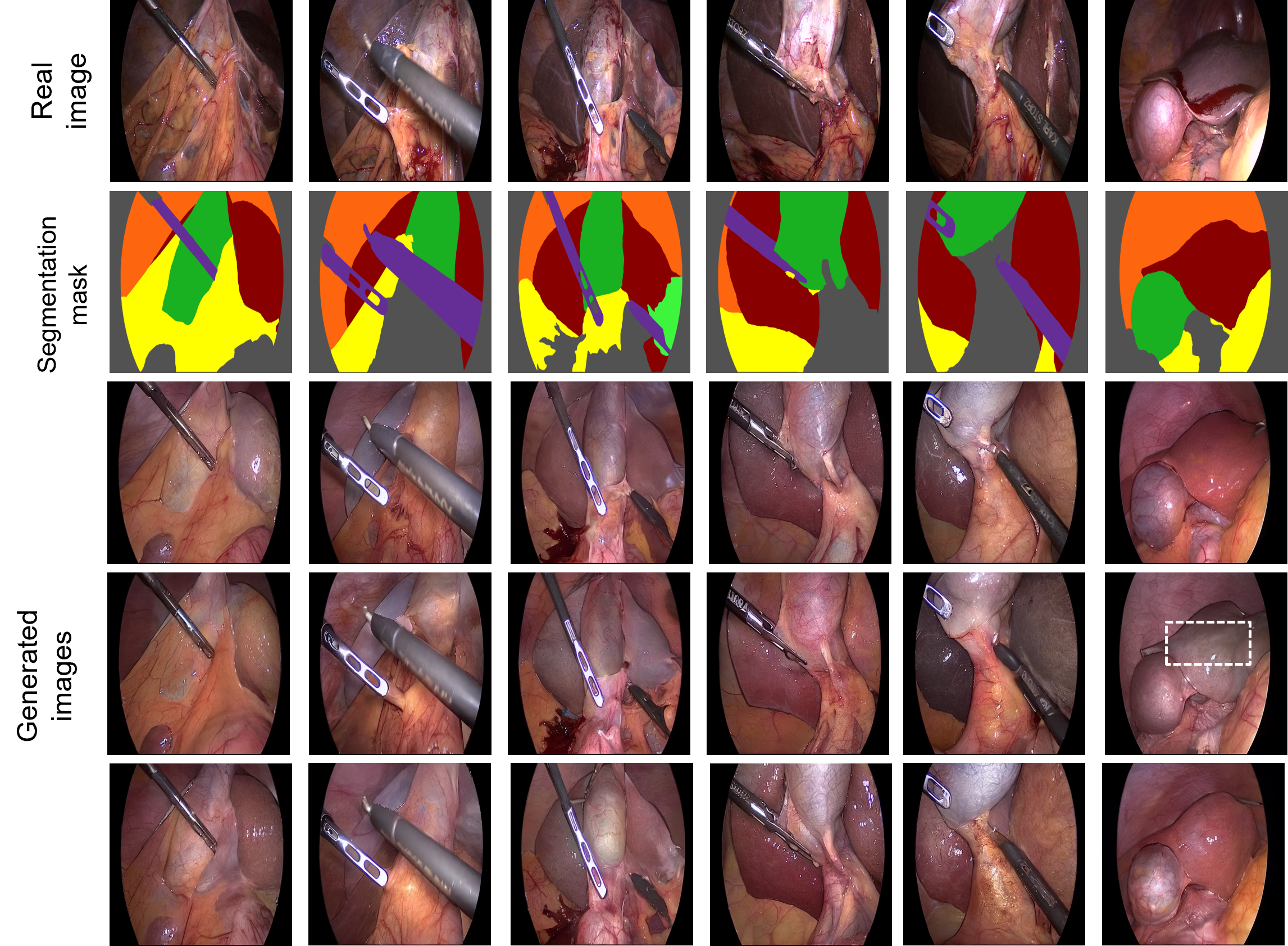}
\end{center}
   \caption{The generated images from the HeiSurf dataset. The real images are shown as representative example, as there exists many different textures of the organs. Our approach is capable of generating different textures for each organ while maintaining the spatial consistency. Our method is capable of generating gall bladders (green color in segmentation mask) which is not completely covered within the fat tissue ($1^{st}$ and $2^{nd}$ column). One failure case is indicated in white box. The texture of the generated liver tissue differs slightly from the real images. The real image in the $6^{th}$ column contains blood on the liver. Our generated images do not synthesize blood pools and could serve effectively as an augmentation method to improve segmentation. Adjusting the CFG scale would be method to rectify this case.}
\label{fig:hs}
\end{figure*}

\begin{figure*}[ht]
\begin{center}
\includegraphics[width=\textwidth]{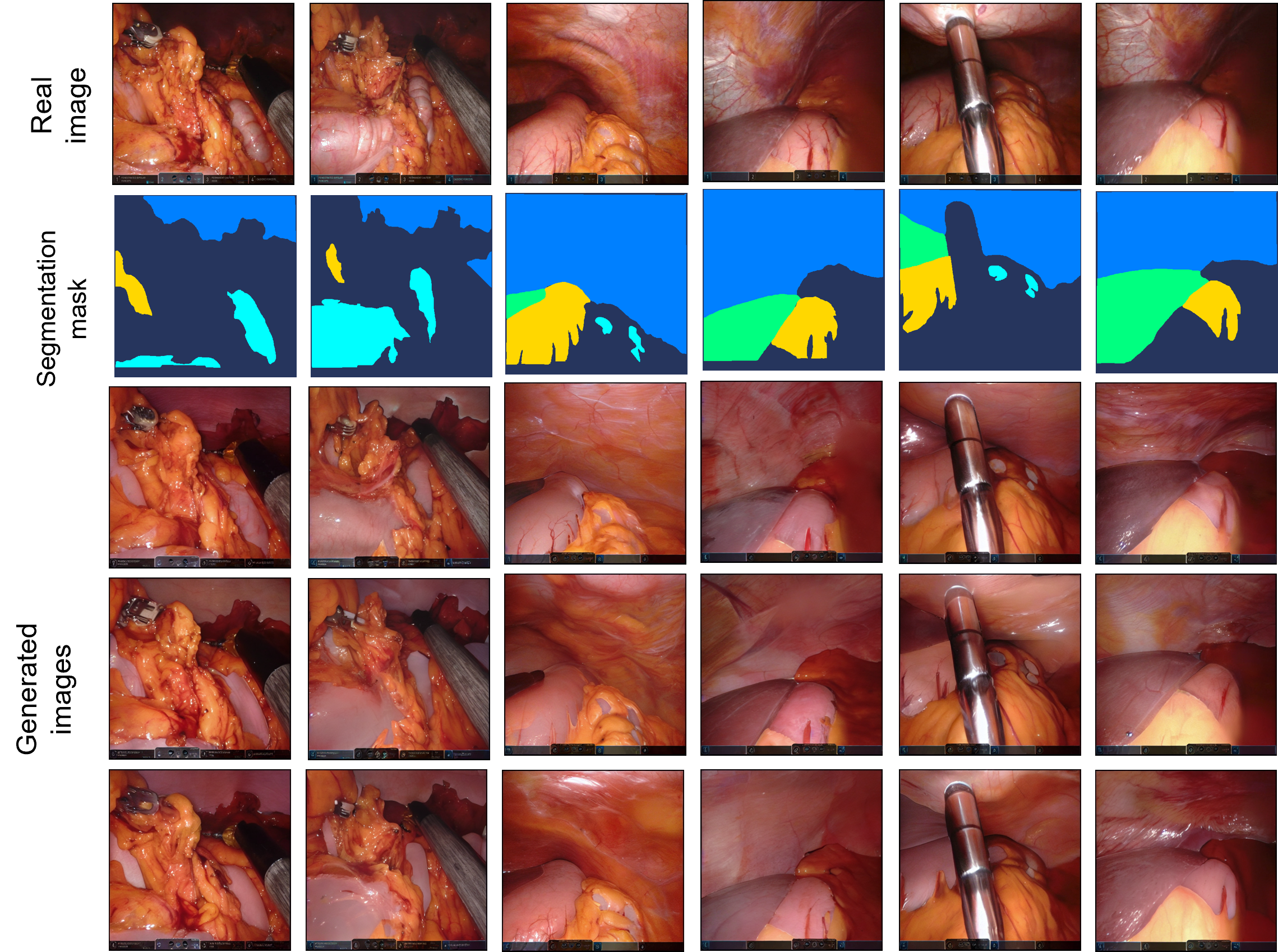}
\end{center}
   \caption{The generated images from the DSAD dataset. The generated images resemble the texture characteristics of the real images. Especially for the liver and stomach (indicated in green and gold color in segmentation mask), we see that the generated images maintain the texture well and adds finer details like vessels (column $3$). It is to be noted that that binary datasets were utilized to generate the organs in this case. This results shows the particular importance of our approach that only real binary datasets can be used to generate multi-class datasets.}
\label{fig:ds}
\end{figure*}

\begin{figure*}[ht]
\begin{center}
\includegraphics[width=\textwidth]{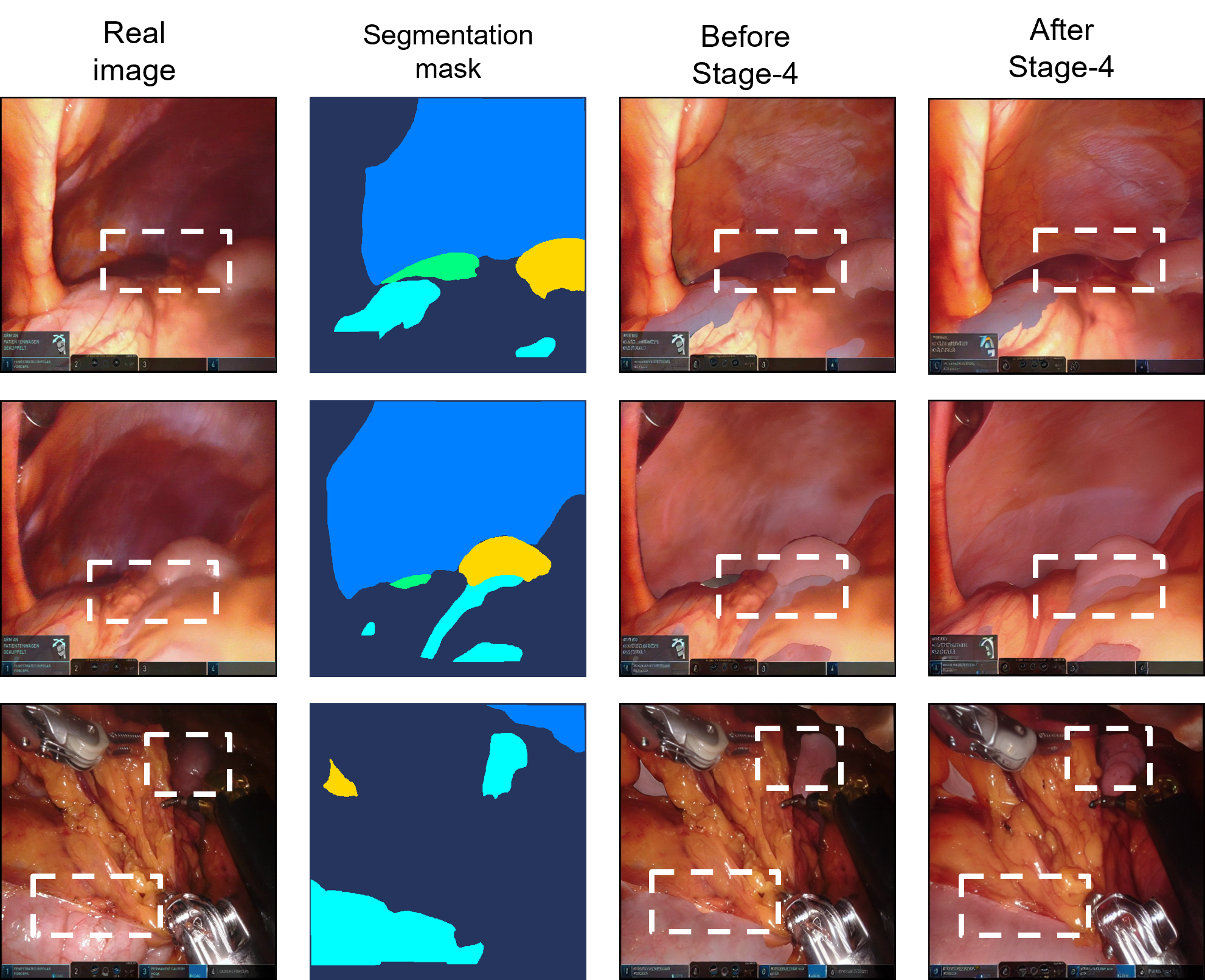}
\end{center}
   \caption{The images generated using the DSAD dataset with and without the Stage-$4$ in our pipeline. The Stage-4 is an image enhancement process that removes the inconsistencies from the image fusion stage. The white boxes indicate the regions comparing the difference between the real image, image after Stage-$3$ ($3^{rd}$ column) and image generated after Stage-$4$. Clearly, fusing the images creates a junction between the different organs. There also exists a slight difference in the background lighting of the generated images from Stage-$3$ ($3^{rd}$ column). To remove them these inconsistencies, the images are processed via a SDEdit method combined with SD model. We use the SD because the model is already aware of the texture of such surgical images. In process leads to a smoother junction between the organs which resembles like the real images.}
\label{fig:ds_in}
\end{figure*}

\begin{figure*}[ht]
\begin{center}
\includegraphics[width=\textwidth]{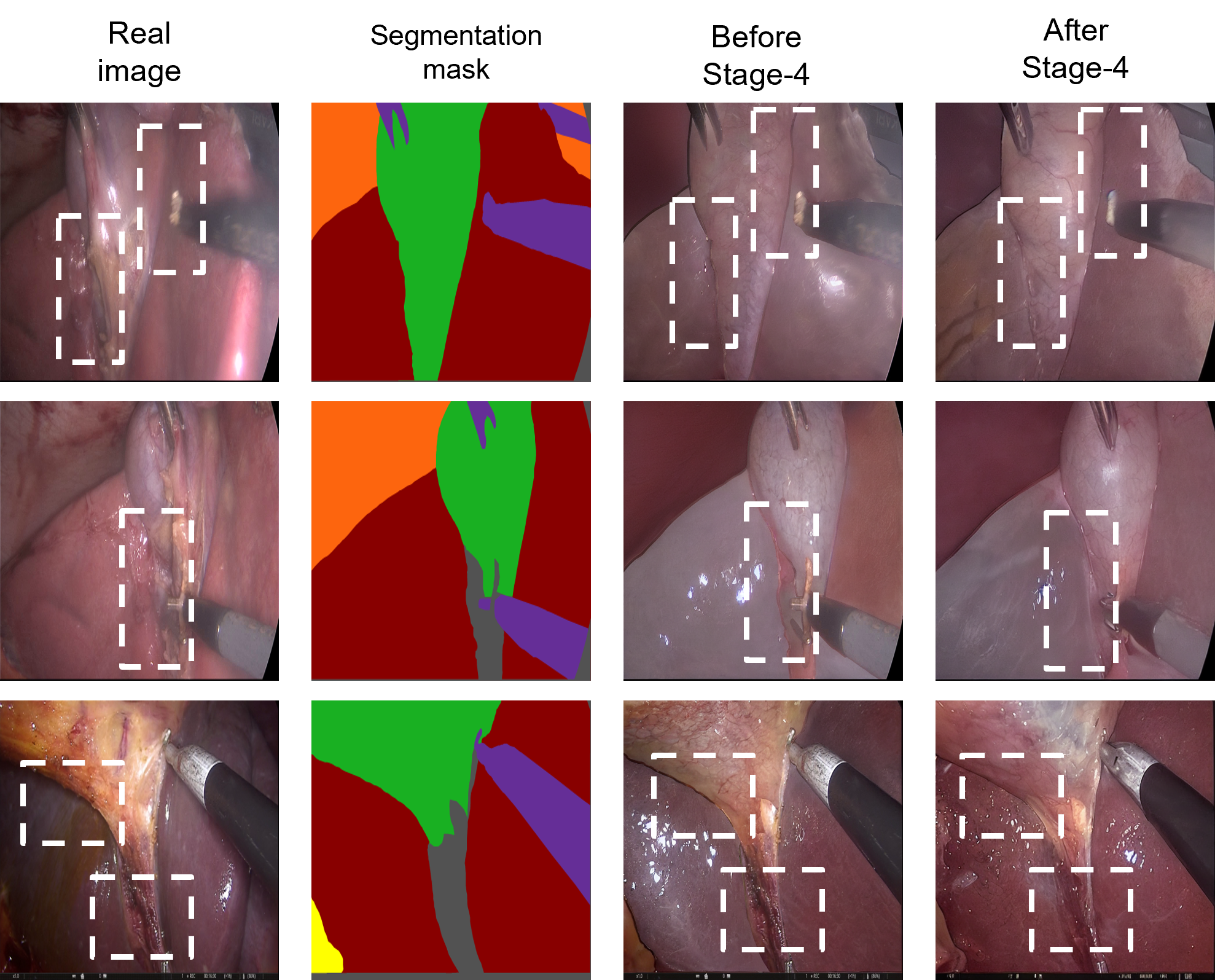}
\end{center}
   \caption{The generated images from the HeiSurf dataset before and after Stage-$3$. In the $1^{st}$ column, we noticed that the images ater Stage-$4$ had finer details like vessels added to the gall bladder. This is advantageous as it enhances the real texture of organs. Additionally, the edges between the organs are smoothened, which is similar to the real images.}
\label{fig:ch_in}
\end{figure*}

\begin{figure*}[ht]
\begin{center}
\includegraphics[width=\textwidth]{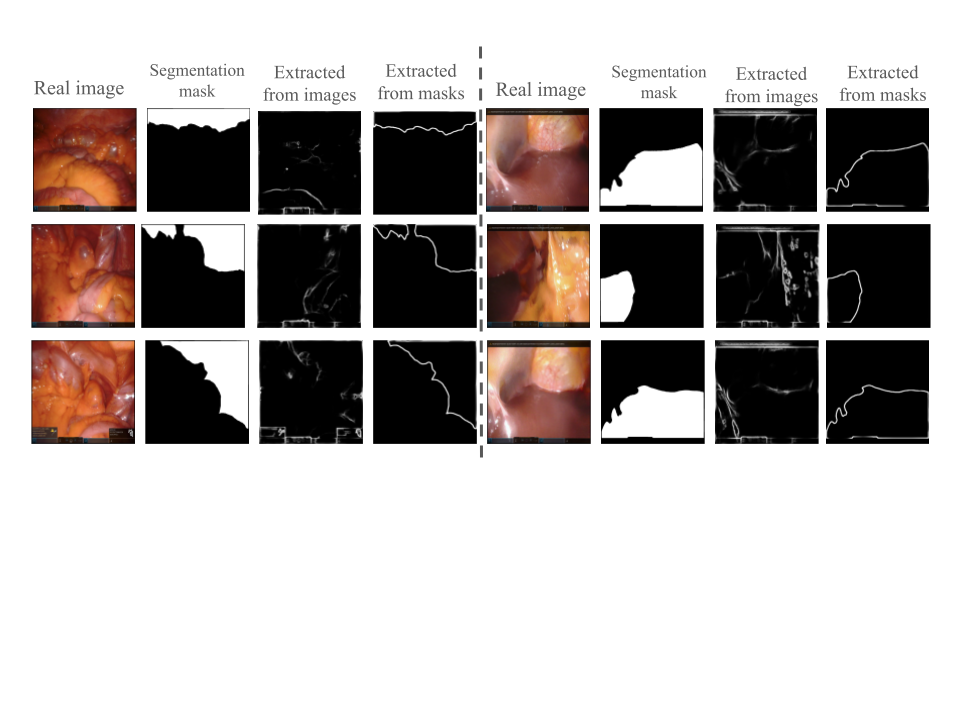}
\end{center}
   \caption{The conditioning signal for the pre-trained ControlNet model in Stage-$2$ of our approach is edge images. Naturally, these edges can be extracted from the real images using a edge detector. However, as shown in column $3$ and $7$, the extracted edges include the tools and other edges which does not correspond to the particular organ. Using such an extracted edge images would lead to inconsistent generation of the organ. Hence, we used the segmentation mask as the input to the extract the edges. As seen in column $4$ and $8$, the extracted edges correspond better to the segmentation mask. In our method, we simultaneously use the same segmentation mask to mask the region for inpainting and also to extract the conditioning signals for the ControlNet.}
\label{fig:map}
\end{figure*}

\begin{figure*}[ht!]
    \centering
    
    \begin{subfigure}{\textwidth}
        \centering
        \includegraphics[width=\linewidth]{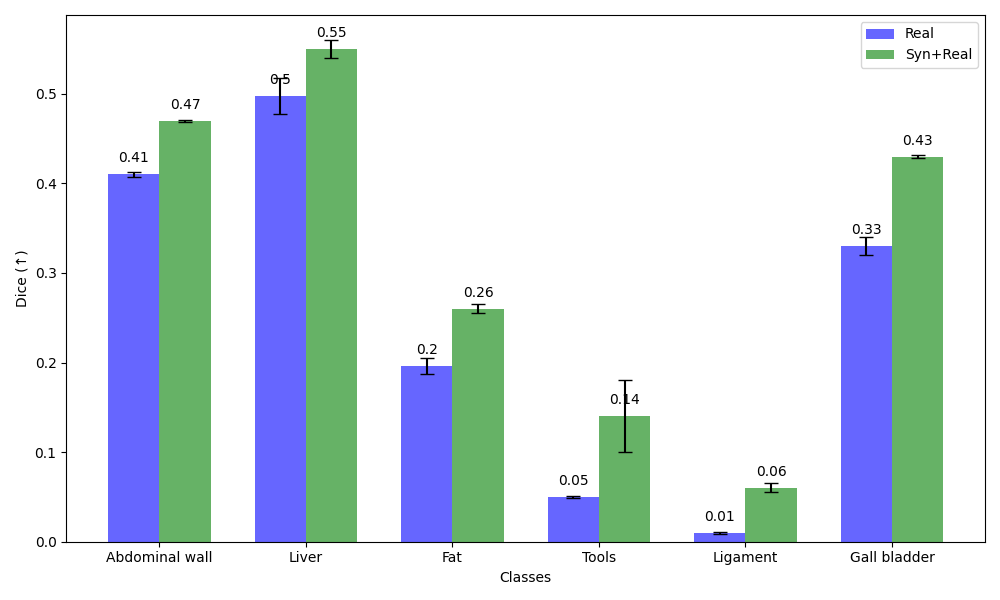}  
        \caption{The dice score on the Cholec80 dataset.}
    \end{subfigure}
    
    \vspace{1em} 
    
    \begin{subfigure}{\textwidth}
        \centering
        \includegraphics[width=\linewidth]{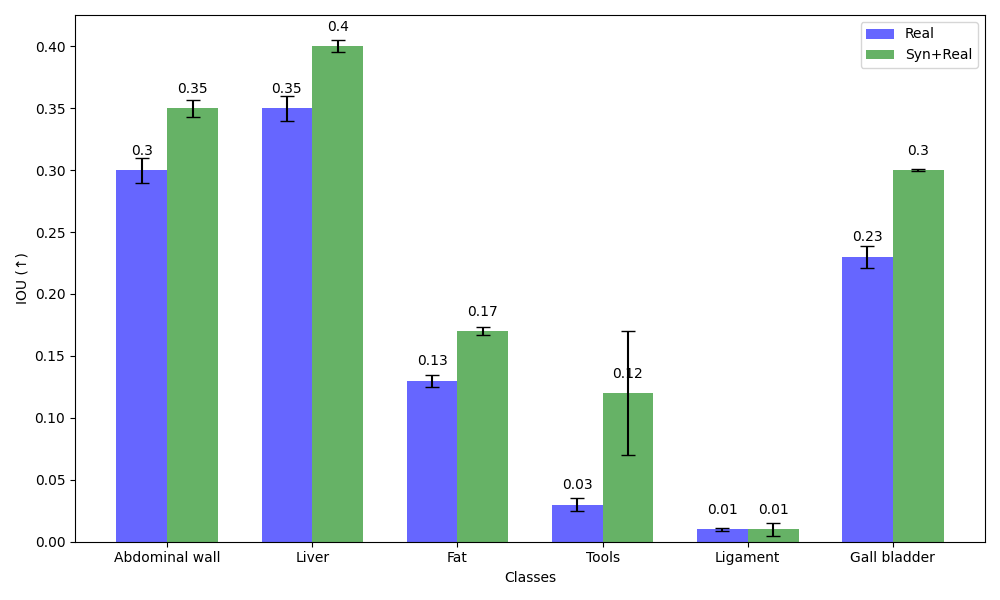}  
        \caption{The IOU score on the Cholec80 dataset.}
    \end{subfigure}

    \caption{The dice and IOU scores for each organ on the Cholec80 dataset. Adding our \emph{Syn} datasets clearly show an imporvement in scores across each organ. Especially, the ligament and gall bladder seems to be segmented particularly well once our \emph{Syn} datasets are added.}
    \label{fig:cholec_combi}
    
\end{figure*}

\begin{figure*}[ht!]
    \centering
    
    \begin{subfigure}{\textwidth}
        \centering
        \includegraphics[width=\linewidth]{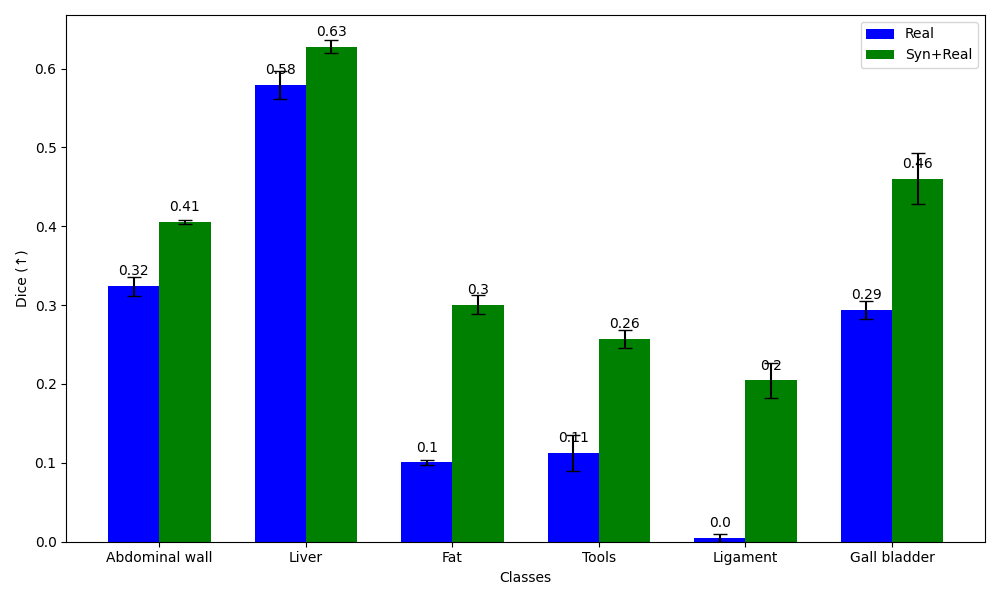}  
        \caption{The dice score on the HeiSurf dataset.}
    \end{subfigure}
    
    \vspace{1em} 
    
    \begin{subfigure}{\textwidth}
        \centering
        \includegraphics[width=\linewidth]{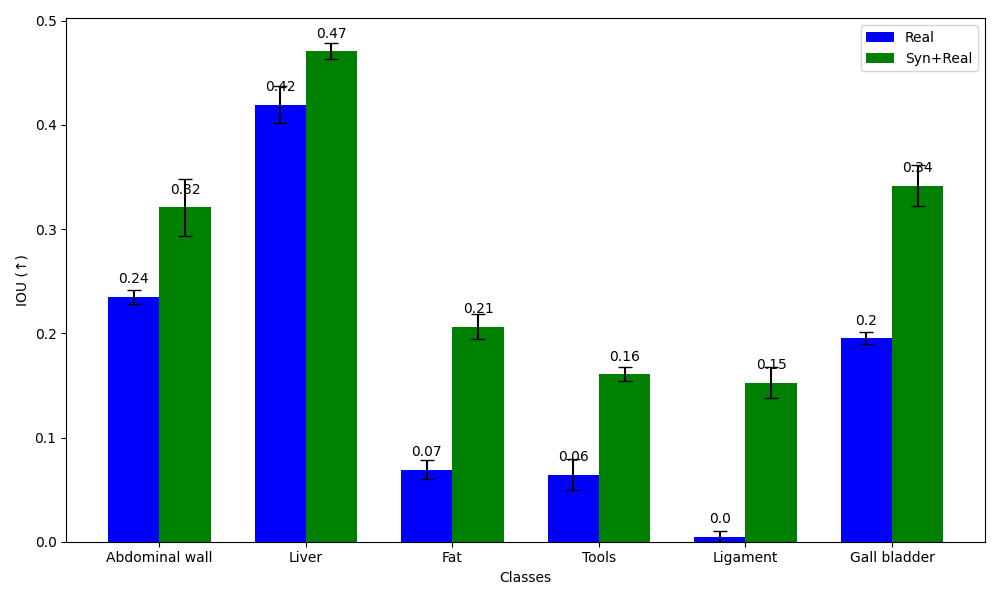}  
        \caption{The IOU score on the HeiSurf dataset.}
    \end{subfigure}

    \caption{The dice and IOU scores for each organ on the HeiSurf dataset. It is evident that combining our \emph{Syn} datasets leads to improved segmentation across six different classes.}
    \label{fig:hs_combi}
    
\end{figure*}

\begin{figure*}[ht!]
    \centering
    
    \begin{subfigure}{\textwidth}
        \centering
        \includegraphics[width=\linewidth]{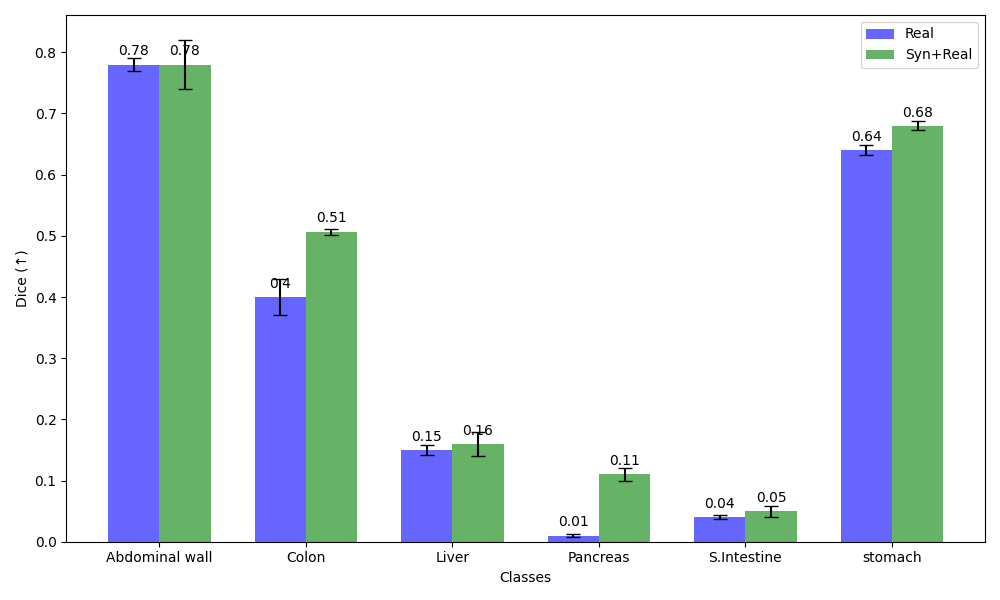}  
        \caption{The dice score on the DSAD dataset.}
    \end{subfigure}
    
    \vspace{1em} 
    
    \begin{subfigure}{\textwidth}
        \centering
        \includegraphics[width=\linewidth]{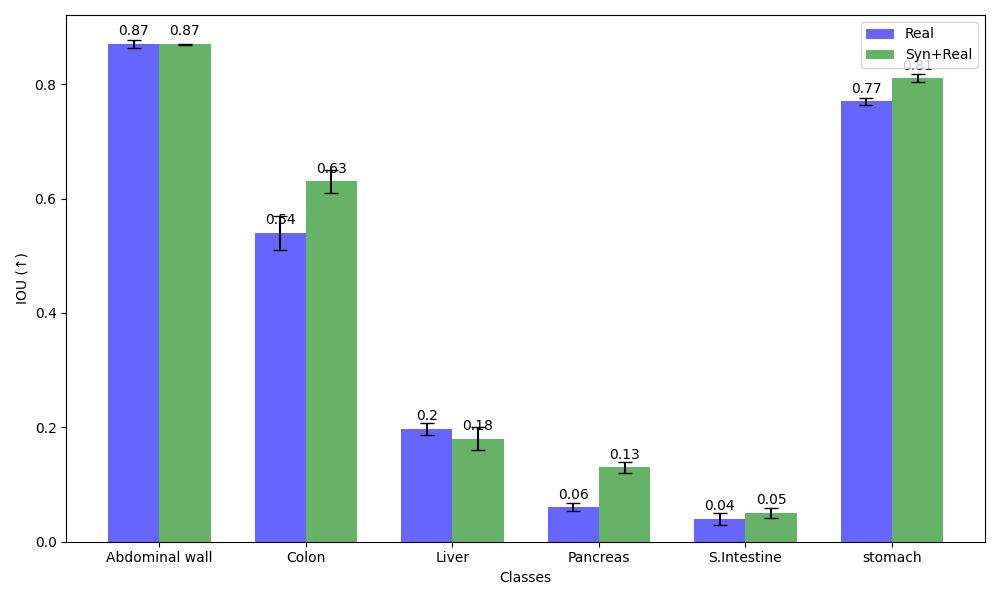}  
        \caption{The IOU score on the DSAD dataset.}
    \end{subfigure}

    \caption{The dice and IOU scores for each organ on the DSAD dataset. We did not notice clear improvements for the abdominal wall and stomach, however, the smaller oragns like the colon, small intestine and pancreas gets segmented better by adding our \emph{Syn} datasets.}
    \label{fig:ds_combi}
    
\end{figure*}

The additional qualitative results from our method on CholecSeg8K, HeiSurf and DSAD datasets are shown in~\cref{fig:ch,fig:hs,fig:ds} respectively. ~\cref{fig:ds_in} and ~\cref{fig:ch_in} show the comparison of images with and without the image enhancement stage. For conditioning the ControlNet we use edge images extracted from the segmentation mask. A comparison is shown in~\cref{fig:map}. The per-organ evaluation scores on the three datasets are shown in~\cref{fig:cholec_combi,fig:hs_combi,fig:ds_combi}. We see consistent improvement across different organs when combining our generated datasets with real images.

\begin{table}
  \begin{center}
    {\small{
    \resizebox{0.75\linewidth}{!}{
\begin{tabular}{ccc|cc}
\toprule
Training method & \multicolumn{2}{c}{CholecT50} & \multicolumn{2}{c}{DSAD} \\
\cline{2-5}
& F1($\uparrow$) & Accuracy($\uparrow$) & F1($\uparrow$) & Accuracy($\uparrow$) \\
\midrule
Real with cl+sp aug. & $0.50_{\pm{0.05}}$ & $0.52_{\pm{0.04}}$ & $0.42_{\pm{0.002}}$ & $0.83_{\pm{0.001}}$ \\
Ours-\emph{Syn}+Real & $\mathbf{0.64_{\pm{0.01}}}$ & $\mathbf{0.63_{\pm{0.02}}}$ & $\mathbf{0.45_{\pm{0.001}}}$ & $\mathbf{0.86_{\pm{0.001}}}$ \\
\bottomrule
\end{tabular}
}
}}
\end{center}
\caption{Surgical target prediction results on two datasets.}
\label{tab:targ}
\end{table}

\begin{table}
  \begin{center}
    {\small{
    \resizebox{0.95\linewidth}{!}{
\begin{tabular}{ccccccccc}
\toprule
Training method & \multicolumn{2}{c}{Scheduler} & \multicolumn{2}{c}{CholecSeg8K} & \multicolumn{2}{c}{HeiSurf} & \multicolumn{2}{c}{DSAD} \\
\cmidrule(lr){2-3} \cmidrule(lr){4-5} \cmidrule(lr){6-7} \cmidrule(lr){8-9}
& DDIM & UniPC & Dice($\uparrow$) & IOU($\uparrow$) & Dice($\uparrow$) & IOU($\uparrow$) & Dice($\uparrow$) & IOU($\uparrow$) \\
\cline{2-9}
\multirow{2}{*}{Only \emph{Syn}} & \checkmark &  & $0.53_{\pm0.01}$ & $0.41_{\pm0.02}$ & $0.35_{\pm0.02}$ & $0.24_{\pm0.01}$  & $0.60_{\pm0.03}$ & $0.51_{\pm0.01}$ \\
& & \checkmark & $0.51_{\pm0.02}$ & $0.39_{\pm0.01}$ & $0.34_{\pm0.01}$ & $0.24_{\pm0.03}$  & $0.58_{\pm0.01}$ & $0.50_{\pm0.01}$ \\ \hdashline
\multirow{2}{*}{Ours-\emph{Syn} + Real} & \checkmark & & $0.68_{\pm0.01}$ & $0.56_{\pm0.01}$ & $0.49_{\pm0.01}$ & $0.36_{\pm0.01}$  & $0.83_{\pm0.01}$ & $0.74_{\pm0.01}$ \\
& & \checkmark & $0.66_{\pm0.02}$ & $0.55_{\pm0.03}$ & $0.49_{\pm0.01}$ & $0.37_{\pm0.01}$  & $0.82_{\pm0.01}$ & $0.74_{\pm0.02}$ \\
\bottomrule
\end{tabular}
}
}}
\end{center}
\caption{Inf. time comparison with different schedulers.}
\label{tab:sched}
\end{table}

\end{document}